\documentclass{article}

\usepackage{tabularx,booktabs,caption}
\usepackage{graphicx}
\usepackage{placeins}
\usepackage{linguex}
\usepackage{enumitem}
\usepackage{natbib}
\usepackage[table, dvipsnames]{xcolor}
\usepackage{arydshln}
\usepackage{caption}
\usepackage{tikz-qtree}
\usepackage{longtable}
\usepackage{amsmath}
\usepackage{float}
\usepackage{amssymb}
\usepackage{bbding}
\usepackage[bottom]{footmisc}
\usepackage{hyperref}
\usepackage{pifont}
\usepackage{booktabs}
\usepackage{multirow}
\usepackage[boxed,ruled,vlined]{algorithm2e}
\usepackage{lipsum}
\usepackage{csquotes}
\usepackage{subcaption}
\usepackage{comment}
\usepackage{endnotes}
\usepackage[left]{lineno}
\usepackage{soul}
\usepackage{rotating}
\usepackage{authblk}
\usepackage[T1]{fontenc}      
\usepackage[utf8]{inputenc}   
\usepackage{setspace}
\usepackage{lmodern}   
\usepackage[htt]{hyphenat}
\usepackage[title]{appendix}
\usepackage{phonrule}
\usepackage[margin=1.2in]{geometry}

\newcommand{\rr}[1]{\textcolor{black}{#1}}
\newcommand{\deltaLoglik}{\Delta \texttt{LogLik}}

\newcounter{globalitem}

\newcommand{\globalitem}{%
  \refstepcounter{globalitem}%
  \item[\theglobalitem.]
}

\title{Back to the Future: The Role of Past and Future Context Predictability in Incremental Language Production}

\author{
  Shiva Upadhye\footnote{Corresponding author: \texttt{shiva.upadhye@uci.edu}} \ and Richard Futrell\\
  {\normalsize\upshape Department of Language Science \\ \normalsize\upshape University of California, Irvine}}
  
\date{}
\begin{document}
\doublespacing
\maketitle
\vspace{-3em}
\begin{abstract}
Contextual predictability shapes how we choose and encode words in production. The effects of a word's predictability given preceding or \emph{past} context are generally well-understood in both production and comprehension, but studies of naturalistic production have also revealed a poorly understood yet robust \emph{backward} predictability effect of a word given only its \emph{future} context, which may be linked to future planning. 
Across two studies of naturalistic speech, we revisit backward predictability using improved operationalizations, introducing \textbf{Conditional Pointwise Mutual Information} (conditional PMI), a conceptually motivated information-theoretic measure that quantifies the information shared between a word and future sequence under the constraints imposed by the past context. Study 1 shows that this measure reproduces predictability effects of future context on phonetic reduction while remaining non-redundant with backward predictability. Study 2 moves from reduction to word choice,  examining substitution errors within a novel generative framework that models lexical contextual influences to predict the identity of the word that surfaces as an error. Within this framework, we find that past-conditioned predictability increases  likelihood of erroneous selection, whereas future-conditioned predictability reduces it. In predicting error identity, conditional PMI emerges as the strongest contextual predictor, subsuming backward predictability. Together, these findings illuminate how past and future context shape word choice and encoding, linking contextual predictability to mechanisms of incremental planning in sentence production.

\end{abstract}

\section{Introduction} \label{sec:intro}
Naturalistic language production shows remarkable variability in \emph{which} words are chosen and \emph{how} they are produced. Much of this variability---particularly in how word-forms are articulated---has been linked to the dynamics of real-time cognitive processing (\emph{cf.} \citealt{bybee2001frequency,pierrehumbert2002word,seyfarth2014word} for offline representation-based accounts). Under this view, an increase in processing difficulty leads to more robust encoding of a word’s phonetic and articulatory detail. Empirically, this relationship is borne out in \emph{probabilistic reduction}: words that are predictable, whether on their own or in context, are more prone to reduction in phonetic detail or word duration (\citealt{balota1989priming,bell2003effects,aylett2004smooth,pluymaekers2005articulatory,bell2009predictability,tily2009syntactic,kurumada2011syntactic,jaeger2017signal,pimentel2021surprisal,ranjan2022linguistic,hashimoto2021probabilistic,wolf2023quantifying} \emph{inter alia}).

Although the effects of contextual predictability on processing difficulty have been studied extensively in psycholinguistic research, much of this work has focused on \textbf{forward predictability}, i.e., the predictability of a word given preceding context. Forward predictability, or its information-theoretic formalization, forward surprisal\footnote{Defined as the negative log probability of the word conditioned on the preceding context, $-\log p(\textrm{word} \mid \textrm{context})$ \citep{shannon1948mathematical}}, has received widespread empirical support as an index of incremental processing difficulty in sentence comprehension \citep{hale2001probabilistic,levy2008expectation,futrell2020lossy,wilcox2020predictive,meister-etal-2021-revisiting,wilcox2023testing,xu-etal-2023-linearity,shainetal24} and production \citep{jurafsky2001evidence,jaeger2006redundancy,bell2009predictability,demberg2012syntactic,mahowald2013info,zhan2018comparing,dammalapati-etal-2019-expectation,harmon2021theory}.

However, a key asymmetry between online language comprehension and production is that the latter affords access to conceptual and linguistic representations beyond the recently produced sequence. Since sentence production involves transforming an abstract conceptual representation of the speaker's message into a linearized linguistic output, the order in which words are planned need not mirror the order in which they are produced \citep{Bock1994LanguageP,ferreira2013syntax,ferreira2018grammatical,Momma2019BeyondLO}. Therefore, when producing an utterance incrementally, speakers have access not only to their \emph{past} output, but also, potentially, to planned or desired \emph{future} representations. 

Consequently, ease of lexical planning may reflect both the facilitative effect of incremental predictability and the constraining influence of upcoming representations (Figure \ref{fig:ProdPlanSchem}). For example, consider the utterances below:

\begin{enumerate}
\globalitem It's not against the law to \underline{\hspace{1cm}} \textbf{alligators across the river} \label{ex:futureContext1}
\globalitem It's not against the law to \underline{\hspace{1cm}} \textbf{alligators through the mail}\footnote{Example utterance from \citet{fromkin2000fromkin}} \label{ex:futureContext2}
\end{enumerate}

\begin{figure*}[t!]
  \centering
      \includegraphics[height=6cm]{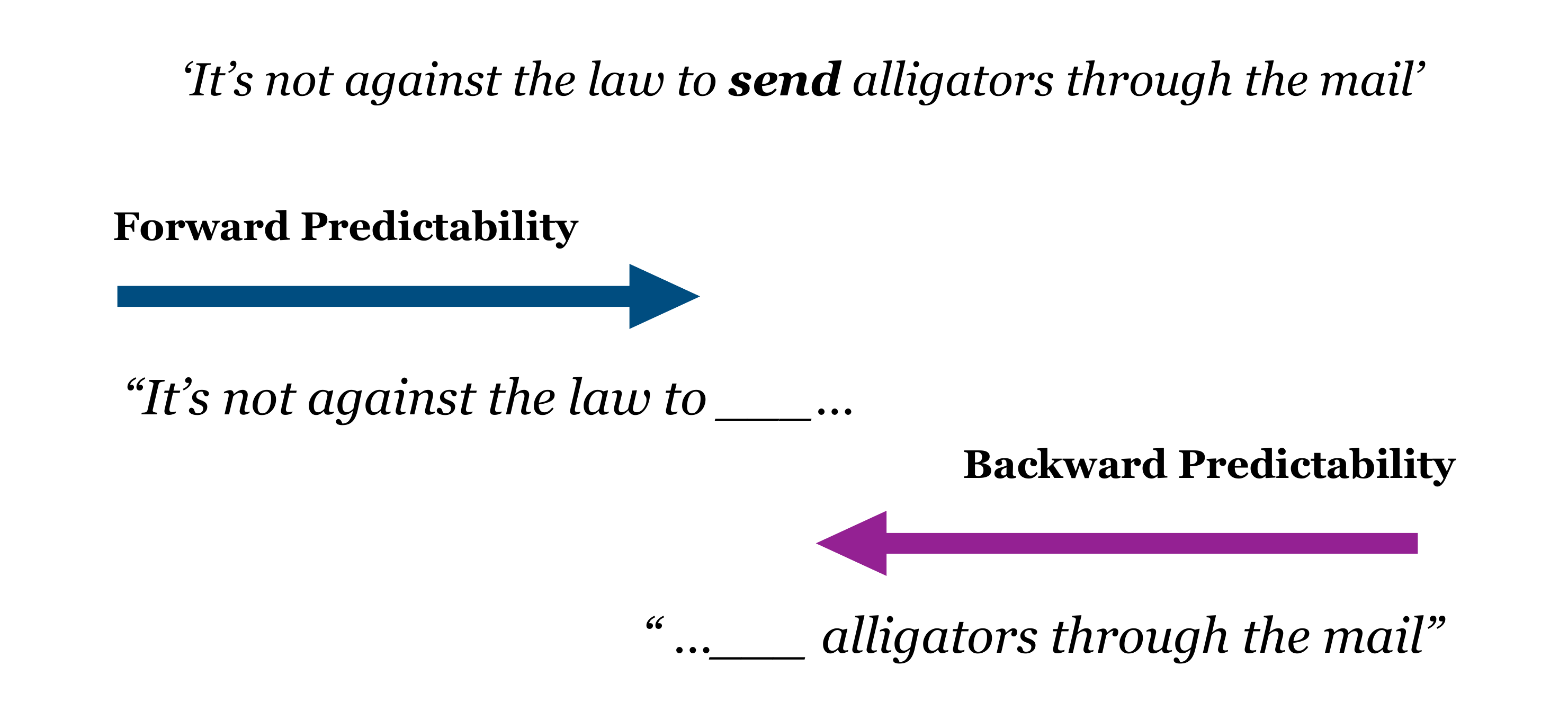}
          \caption{An illustration of forward and backward-looking contextual predictability effects in naturalistic production.}
\label{fig:ProdPlanSchem}
\end{figure*}

The shared preceding context licenses several compatible verbs at position $t=7$ (e.g., \emph{drink}, \emph{dream}, \emph{gamble}, \emph{jaywalk}, \emph{dance} etc.) while the following context (\textbf{bolded}) in both utterances narrows down word choice to transitive verbs that can take \emph{alligators} as a direct object. Moreover, the following context in (\ref{ex:futureContext2}) imposes a much stronger constraint on word choice: for example, even though the verb \emph{send} can appear in both contexts, it is more predictable given `through the mail' than `across the river.' This backward-looking effect of upcoming material has been broadly referred to as \textbf{backward predictability}, which captures how predictable a word is under an upcoming or future sequence. A number of studies have shown that backward predictability emerges as a robust predictor of planning difficulty in naturalistic speech, as reflected in both acoustic enhancement \citep{pluymaekers2005articulatory,bell2009predictability,hashimoto2021probabilistic,ranjan2022linguistic} and the prevalence of disfluencies \citep{goldmaneisler1958speech,shriberg1996disfluencies,dammalapati-etal-2019-expectation,dammalapati-etal-2021-effects,harmon2021theory}, even after controlling for the effects of other probabilistic variables such as lexical frequency and forward predictability. 

Prior work has also shown that predictability from the past context (i.e., forward predictability) and future context (i.e., backward predictability) show asymmetric effects on word durations. Notably, \citet{bell2009predictability} observed that the effects of contextual predictability appeared to be modulated by frequency for function words: while durations of high-frequency function words were sensitive only to predictability from the past, durations of mid-to-low frequency words were affected by predictability from the following word. By contrast, content word durations were sensitive only to predictability from the following but not the preceding word (\emph{cf.} \citealp{ranjan2022linguistic}), potentially reflecting stronger associations between content words and upcoming material.

\begin{figure*}[t!]
  \centering
      \includegraphics[width=\textwidth]{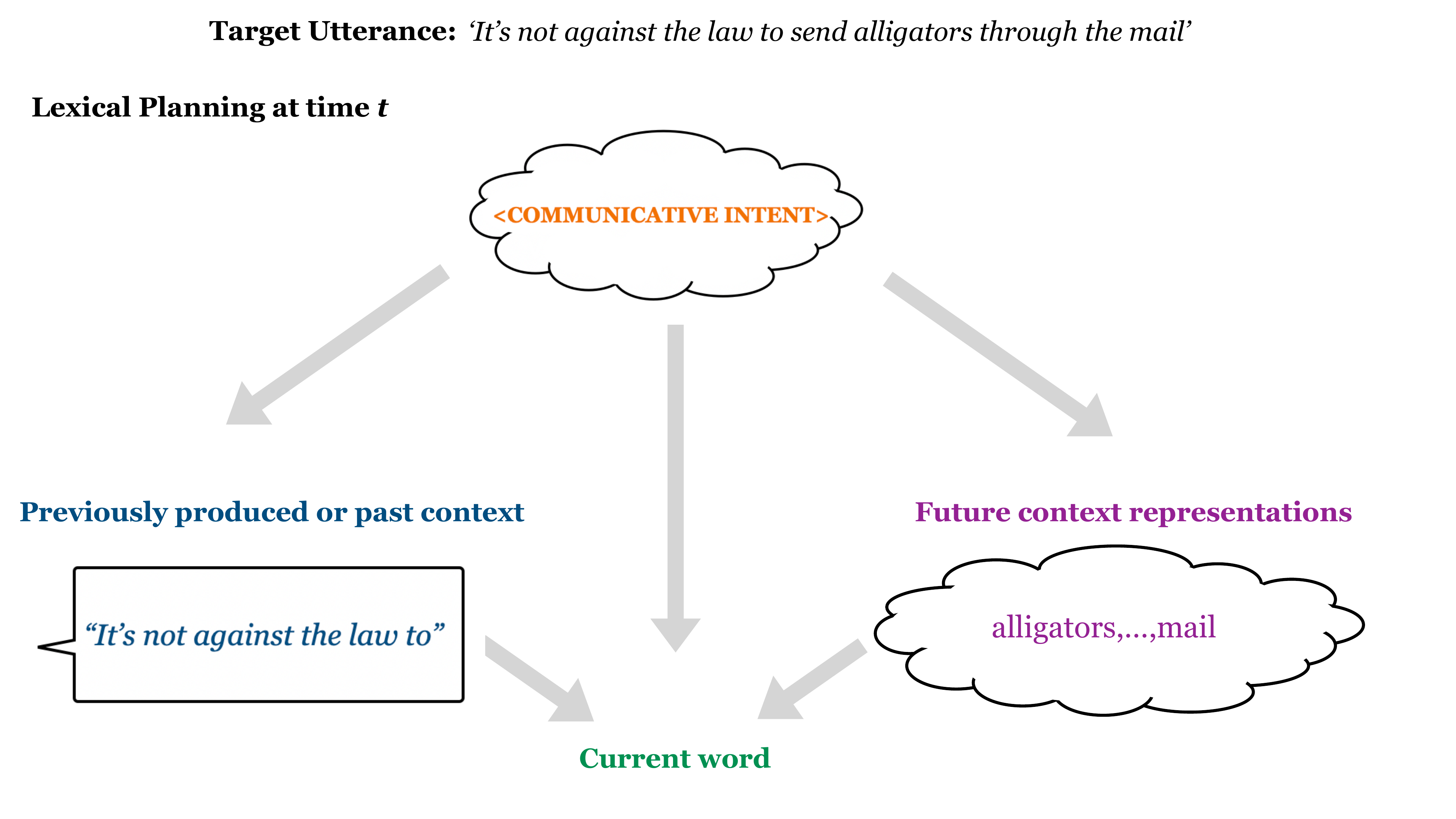}
          \caption{Effects of communicative intent and context-based information sources on lexical planning in incremental language production.}
\label{fig:ProdPlanSchem}
\end{figure*}

A number of conceptual interpretations of backward predictability effects have been proposed. In the articulatory-motor domain, \citet{pluymaekers2005articulatory} propose a link between the backward predictability effect and anticipatory motor planning: when the current and following word are highly informative about each other, speakers speed-through the articulation of the current word to initiate motor planning of the upcoming word, resulting in reduced durations. This account is \emph{prospective} in that it attributes (non-)reduction to the availability of upcoming material. In contrast, \citet{bell2009predictability} put forth a \emph{retrospective} explanation of this effect: predictability---whether unigram, forward or backward---facilitates retrieval of the current word, leading to reduced buffering and smoother coordination between planning and articulation. Another interpretation of this quantity comes from \citet{harmon2021theory}, who treat the predictability of the preceding sequence conditioned on the present word as form of \emph{reactivation}. Once reactivated (through repetition), the preceding sequence serves as a cue that facilitates retrieval of the upcoming word. 

One barrier to finding a clear interpretation of the backward predictability effect is that we do not yet have the full empirical picture of the effects of predictability on word duration and choice. A key limitation of previous work is that the commonly-used measure of backward predictability assumes implicitly that the future context is produced \emph{independently} of the previously produced or past context. That is, although the measure captures the predictive effects of future material on the present word, it disregards the specific context in which the speaker produces that future material. 

Crucially, such an assumption is inconsistent with our understanding of the architecture underlying sentence production. Incremental production begins with conceptual, order-agnostic representations of the speaker's message, which are subsequently \emph{serialized} to satisfy both typological and articulatory-motor constraints \citep{lashley1951problem,levelt1981speaker,Bock1994LanguageP}. Therefore, the \emph{past} and \emph{future} relative to a notional \emph{present} word may not be intrinsic to how speakers plan their utterance; rather, they reflect a property of how the overt production of the observed utterance unfolds over time. Since both are jointly constrained by the speaker's message, past and future context can be viewed as observable approximations of it.

Moreover, cross-linguistic evidence shows that production planning can proceed \emph{non-linearly}, driven by structural relations between words that may be positioned arbitrarily far apart in the serialized form of the utterance \citep{schriefers1998producing,lee2013ways,momma2016timing,sauppe2017word,Momma2019BeyondLO}. These findings suggest that previously produced and planned yet unrealized future words may be directly linked via relations that are not mediated by the choice of the current word. 

However, since backward predictability measures only the predictability of the current word from the future context and ignores the past context altogether, it cannot capture such direct associations between words in the past and future. Beyond this conceptual limitation, the interpretation of backward predictability in regression models that also include forward predictability is complicated by the fact that the two probabilistic variables are highly collinear \citep{bell2009predictability}, which obfuscates the distinct predictive contributions of past and future context. 

Here, we give a more complete characterization of the predictive effects of past and future context on speaker choices in naturalistic production. Conceptually, we argue that backward predictability reflects one particular operationalization of future-context predictability---one that assumes independent effects of the past and future contexts. We propose an alternate mathematical operationalization of future context effects based on the \textbf{Conditional Pointwise Mutual Information (PMI)} \citep{fano1961transmission} of the current word and future after extracting information from the past context---a measure that avoids the independence assumption implicit in backward predictability. Methodologically, we address concerns about collinearity and language modeling to improve the estimation of contextual predictability measures. 

Empirically, this work is grounded in two studies that illuminate the nuanced effects of contextual predictability on lexical planning. Our first study revisits \emph{probabilistic reduction} as a testbed for evaluating the improved contextual measures, including our proposed conditional PMI-based formulation of future context predictability. Our second study investigates the effects of context on word choice by independently modeling how lexical availability, communicative alignment, and past- and future-context predictability shape the content of lexical substitutions in naturalistic productions.

Our findings have implications for accounts of probabilistic reduction and incremental planning in sentence production. Through our first study, we demonstrate that our principled conditional PMI-based measure qualitatively replicates the effects of backward predictability in predicting word durations while also explaining non-redundant variance in duration. However, it does not reproduce the asymmetric effects of context on function and content words observed by \citealt{bell2009predictability}. Instead, we find that lexical and contextual predictability from either direction predict reduction for both content and function words, although function word durations exhibit reduced sensitivity to predictability overall. Finally, our model of substitution errors broadens the empirical scope of research on trade-offs in word choice, which has so far been examined largely independently of the informational constraints imposed by sentential context. This analysis reveals distinct functional pressures imposed by past and future context on the choice of substitution errors, corresponding respectively to goal-invariant (i.e., availability-based) and goal-directed mechanisms in sentence planning.

\section{Proposed Framework} \label{sec:modeling}

\begin{figure*}[t!]
  \centering
      \includegraphics[width=\textwidth]{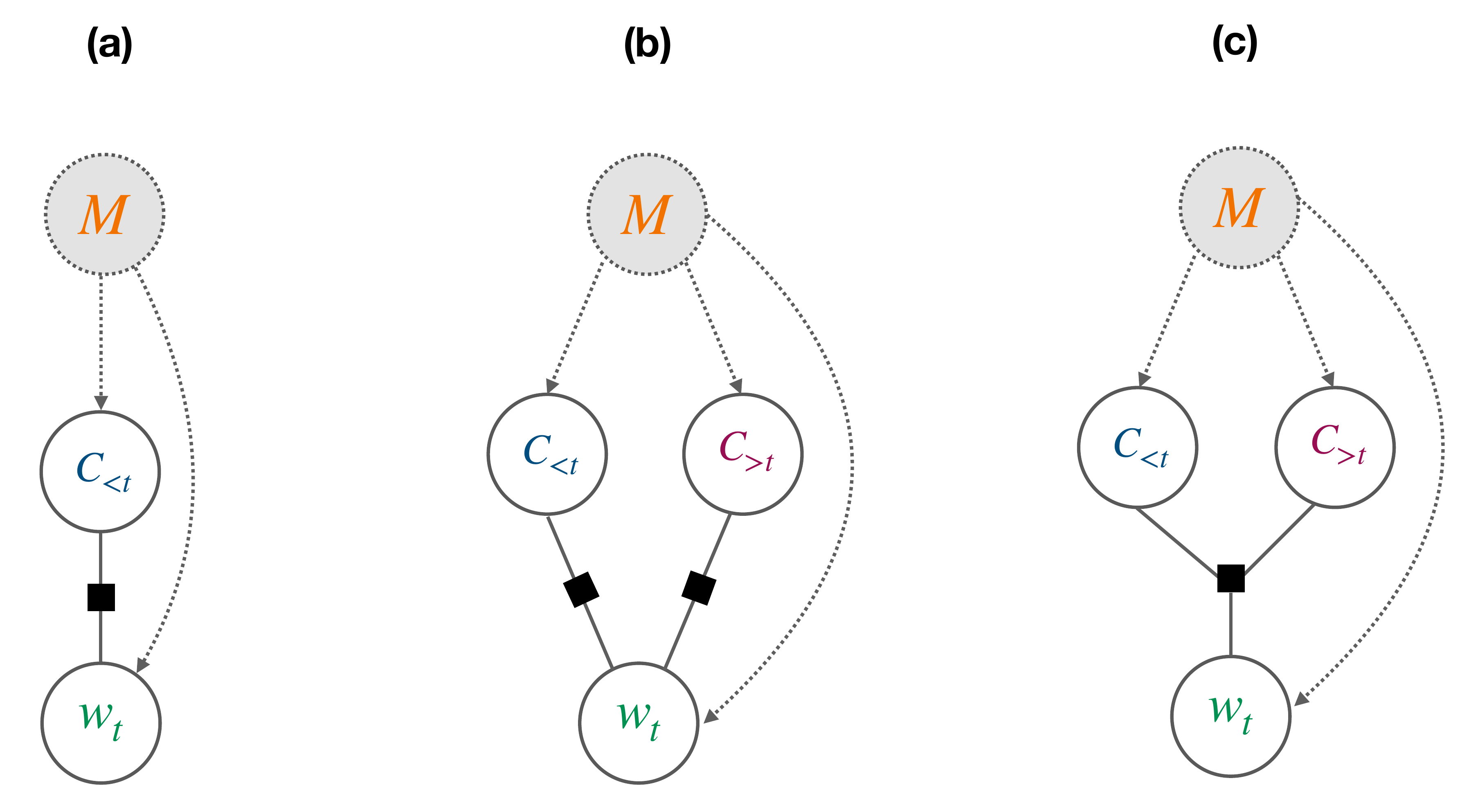}
          \caption{An illustration of the information-processing dependencies between the speaker's message ($M$), past context ($C_{<t}$), future context ($C_{>t}$), and the current word ($w_t$). Here, we treat $M$ as a latent variable, and $C_{>t}$ as an observed variable, even though for the speaker, the future is not realized until after the production of $w_t$. Solid lines indicate explicit conditioning dependencies while dashed lines indicate causal influences between the latent variable $M$ and the contextual representations. Black squares indicate \textbf{factors}, which define functions between connected variables. For example, the black square in (c) denotes that $w_t$ is a function of both $C_{<t}$ and $C_{>t}$, whereas in (b), $w_t$ is determined by  independent functions of $C_{<t}$ and $C_{>t}$.  Left \textbf{(a)}: Current word depends solely on past context, as reflected in forward predictability. Middle \textbf{(b)}: the current word is influenced separately by past and future context---the assumption implicit in backward predictability. Right \textbf{(c)}: Current word is jointly influenced by the past and future context, as in the case of bidirectional word probability.}
\label{fig:dep-graph}
\end{figure*}

The goal of the present study is to estimate the distinct contributions of contextual predictability from the past and future on articulatory duration and word choice in naturalistic production. For the speaker, when future representations are entirely unavailable, the current word may be predicted solely from its past context (Figure \ref{fig:dep-graph}a), as quantified by forward predictability:

\begin{equation}
    p(w_t \mid C_{<t}) \label{eq:fwPred}
\end{equation}

\noindent where $w_t$ is a word at time $t$ and $C_{<t} = w_1 \cdots w_{t-1}$ is the previously produced or past context. Backward predictability provides an analog of this effect in the reversed direction:

\begin{equation}
    p(w_t \mid C_{>t}) \label{eq:bwPred}
\end{equation}

\noindent where $C_{>t} = w_{t+1} \cdots w_{N}$ is an upcoming sequence of words, with $w_{N}$ serving as the last word of the sentence. This measure can be seen as quantifying the predictability of the current word under a future sequence, for a speaker who has more or less fully planned the future context as it will be produced.  

As in some previous work (e.g., \citealt{pluymaekers2005articulatory}), backward associations between the current word and the future word or sequence can also be modeled using (unconditional) Pointwise Mutual Information or PMI:

\begin{equation}
    \textrm{PMI}(w_t ; C_{>t}) = \log p(w_t \mid C_{>t}) - \log p(w_t) \label{eq:uncond-pmi}
\end{equation}

\noindent where $\textrm{PMI}(w_t ; C_{>t})$ measures the strength of association between the word $w_t$ and the future $C_{>t}$ after accounting for how frequently they occur independently. Thus, a positive value of $\texttt{PMI}(w_t ; C_{>t})$ indicates that the word and the future sequence are highly informative about each other i.e., are more likely to co-occur than would be predicted by chance alone. 

However, when viewed as an index of future planning difficulty, backward predictability, or its information-theoretic analog $\textrm{PMI}(w_t ; C_{>t})$, is limited in one crucial respect: it treats the speaker's current state---i.e., the context produced so far---as irrelevant to upcoming production. Formally, it models the conditional dependence of the current word ($w_t$) on the future context ($C_{>t}$), but disregards the (past) context ($C_{<t}$) in which both are anchored. In effect, this treats past and future context as independent sources of information about $w_t$ (Figure~\ref{fig:dep-graph}b)---an assumption that does not hold, since both are ultimately shaped by the speaker's intended message ($M$). Instead, if the past and future jointly constrain the current word (Figure~\ref{fig:dep-graph}c), then predictability from the future alone may over- or underestimate how \emph{uniquely} informative that future context is. 

\subsection{Quantifying Future Context Predictability} \label{sec:quantifying-FCP}
In this work, we relax the assumption that past and future context have independent predictive effects on lexical planning. With this goal in mind, we propose a principled alternative to backward predictability, based on the \textbf{conditional PMI} of the current word $w_t$ and the future context $C_{>t}$ given the past context $C_{<t}$:

\begin{align}
    \textrm{conditional PMI}(w_t; \ C_{>t} \mid C_{<t}) &= \log \frac{p(w_t \mid C_{>t}, C_{<t})}{p(w_t \mid C_{<t})} \\
    & = \underbrace{\log p(w_t \mid C_{>t}, C_{<t})}_{\textrm{Bidirectional word probability}} - \underbrace{\log p(w_t \mid C_{<t})}_{\textrm{Forward word probability}}
    \label{eq:PMI}
\end{align}

The bidirectional probability term in Eq.~\ref{eq:PMI} quantifies how past and future context \emph{jointly} predict the current word, thereby preserving any associations between non-adjacent words within the utterance. Subtracting the forward probability then isolates how much \emph{more informative} the future context is about the current word, given the speaker's current state, i.e., the context they have already produced. It is this second step that distinguishes our measure from prior work that uses either vanilla backward probability or bidirectional probability, neither of which isolates the unique contribution of future context predictability once the past has been taken into account. \citet{wolf2023quantifying}, for instance, show that bidirectional context shares higher mutual information with word duration than past context alone; however, because their measure does not extract out the past, it cannot establish how much of that gain is attributable to future context specifically. 

In comparison, backward predictability provides an estimate of future context predictability that does not take into account the already produced context and can therefore be considered \emph{context-independent} across all possible past sequences that the speaker could have produced\footnote{Put differently, backward predictability can be viewed as analogous to the bidirectional probability in Eq. \ref{eq:PMI} but with the past context marginalized out i.e., $\sum\limits_{c} p(w \mid C_{>t}, C_{<t} = c) \cdot p(C_{<t} = c)$}. Cognitively, it corresponds to a state of epistemic uncertainty in which the speaker aggregates over all possible past sequences: that is, a state where the realized past itself is treated as no more privileged than the alternatives.

A positive value of conditional PMI\ suggests that the current word and future sequence are informative about each other \emph{beyond} what can be predicted by the past context. For instance, consider the following sentences:

\begin{enumerate}
\globalitem It's not against the law to \textbf{[send]} alligators through the mail 
\label{ex:positive-pmi-ex}
\globalitem She poured the \textbf{[gin]} into her laptop \label{ex:negative-pmi-ex}
\end{enumerate}

Ex. (\ref{ex:positive-pmi-ex}) provides an example of positive conditional PMI: the current word (\emph{send}) becomes more predictable in bidirectional context, which includes the informative upcoming sequence \emph{through the mail}. However, (\ref{ex:negative-pmi-ex}) illustrates an instance where conditional PMI\ may be negative, as the current word (\emph{gin}) is highly predictable under the past context, but becomes less predictable once the future is taken into account. 

\begin{figure*}[t!]
  \centering
      \includegraphics[width=\textwidth]{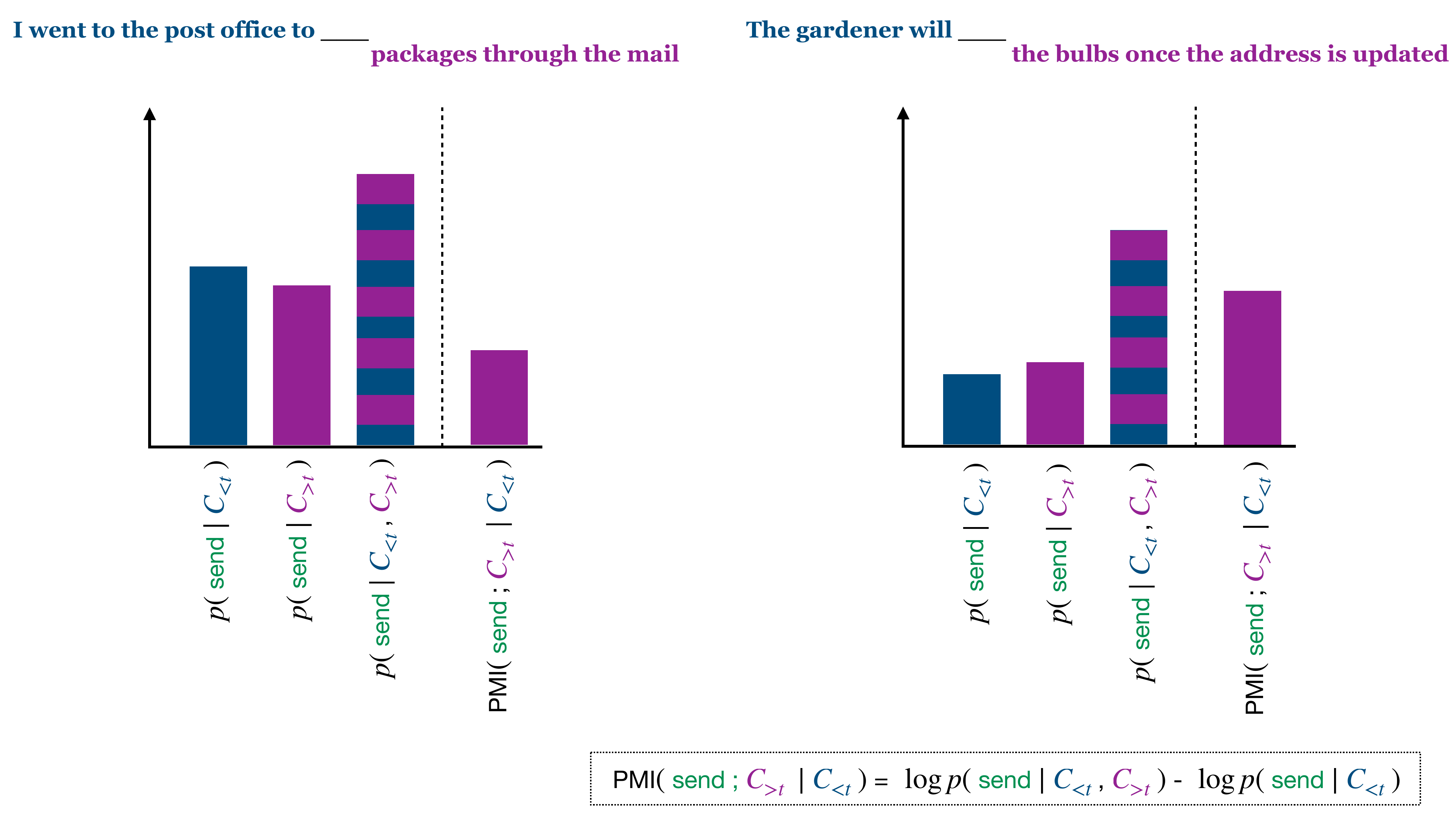}
          \caption{Simulated comparison of forward probability $p(w_t \mid C_{<t})$ (in blue), backward probability $p(w_t \mid C_{>t})$ (in purple), bidirectional probability $p(w_t \mid C_{<t}, C_{>t})$ (striped blue and purple), and conditional \textbf{PMI}($w_t \ ; \ C_{>t} \mid C_{<t}$) (in purple) to illustrate when backward predictability and conditional PMI diverge in their estimations of the predictive contribution of future context, with a focus on how backward predictability may over- or underestimate this contribution. Recall that conditional PMI is defined as the difference between log-transformed bidirectional and forward probabilities. The y-axis is illustrative and reflects only the relative magnitude of each quantity; it should be noted, however, that the forward, backward, and bidirectional probabilities are probabilities, whereas conditional PMI is in nats.}
\label{fig:measure-comparison}
\end{figure*}

\rr{Moreover, as noted earlier, by explicitly controlling for the contribution of the past context, conditional PMI more accurately captures the unique information gained from the future context (Figure \ref{fig:measure-comparison}). For example, consider the sentence  below:}

\begin{enumerate}
\globalitem I went to the post office to \textbf{[send]} packages through the mail 
\label{ex:diff-pmi-bwpred-over}
\end{enumerate}

\rr{Unlike (\ref{ex:positive-pmi-ex}), where the past context is not predictive of the target \emph{send}, (\ref{ex:diff-pmi-bwpred-over}) illustrates an instance where the preceding sequence ``I went to the post office to'' strongly predicts \emph{send} such that the \emph{additional} information gained from the future context is limited. Because backward predictability quantifies the predictive contribution of the future independent of the past context, it may, in this case, overestimate how uniquely informative the future context ``packages through the mail'' is about the target. As noted earlier,  conditional PMI is defined as the difference between log-transformed bidirectional and forward probabilities; consequently, when both of these probabilities are high, their difference is low, reflecting a limited predictive contribution of the future context. Crucially, this cannot be captured by backward predictability, whose value depends solely on whether the predictability from the future context is high.}

\rr{Similarly, backward predictability may also underestimate the predictive influence of the future context. Consider the example below:}

\begin{enumerate}
\globalitem The gardener will \textbf{[send]} the bulbs as soon as the customer's address is updated. 
\label{ex:diff-pmi-bwpred-under}
\end{enumerate}

\rr{In (\ref{ex:diff-pmi-bwpred-under}), the past context does not strongly predict the target \emph{send} and instead favors verbs such as \emph{plant}, \emph{grow}, or \emph{prune}. Given future context alone, \emph{send} or \emph{ship} may become moderately predictable, such that the backward predictability of \emph{send} may exceed its forward predictability. Yet, the future context is ambiguous in isolation: since \emph{bulbs} also admits the \emph{electric bulb} sense of the word, the future also licenses verbs such as \emph{repair}, \emph{change}, or \emph{install}. This ambiguous reading of the future context is resolved when considered in combination with the past context. Although the past  is less constraining on its own, it provides the disambiguating cue \emph{gardener}, which rules out candidates such as \emph{repair} or \emph{install}, while verbs such as \emph{send} or \emph{ship} still remain more compatible with the future context. The future is therefore more informative about the target when anchored in the past than backward predictability reveals in this particular example.}

Beyond this, a key mathematical property of Pointwise Mutual Information is that it is symmetric: it does not commit to whether the future sequence influences the choice of the current word or vice versa. A positive value of conditional PMI\ is, therefore, compatible with (i) a \emph{retrospective} interpretation, i.e., the planned future sequence is highly informative about word choice at the current time-step, or (ii) a \emph{prospective} interpretation, wherein the choice of the current word facilitates production of upcoming material. 

This latter interpretation is broadly aligned with the concept of \emph{value-to-go} \citep{sutton1998reinforcement,todorov2009efficient} in action planning, optimal control, and reinforcement learning. Value-to-go is the total expected value of taking an action, consisting of the immediate utility of the action itself and the expected value of future actions after it. In theories of optimal planning, an agent chooses actions to maximize this value-to-go function. As applied to language production, value-to-go includes predictability as a factor---more predictable actions or words are higher value because they incur less cognitive cost, and can be produced in a more routinized way \citep{gershman2020rationally,gershman2020origin,lai2021policy,futrell2023information}. Therefore, an optimal planning agent will tend to produce a  word $w_t$ that increases the predictability of a high-value \emph{future} sequence $C_{>t}$, as measured by the surprisal
\begin{equation}
\label{eq:cond-future-surprisal}
-\log p(C_{>t} \mid w_t, C_{<t}).
\end{equation}
We show how conditional PMI\ can be derived from Eq.~\ref{eq:cond-future-surprisal} in Appendix \ref{app:derivation-PMI}.

To enable a principled comparison between conditional PMI and backward predictability, we must account for the fact that backward predictability is itself correlated with forward predictability, whereas conditional PMI partials out the standalone contribution of the past, effectively producing a decorrelated quantity. We therefore evaluate conditional PMI against the following decorrelated alternative to backward predictability\footnote{A commonly used alternative method for de-correlating predictors is \textbf{residualization}---a process that involves regressing the variable of interest on a correlated variable and using the residuals or prediction errors in lieu of the target variable. While residualization can effectively de-correlate predictors by removing covariance between these variables, this process may alter the interpretability of the construct represented by the target variable \citep{breaugh2006rethinking,wurm2014residualizing}. In other words, a residualized variant of backward predictability may address the problem of collinearity, but may render this effect less cognitively interpretable. See also \citet{opedal-etal-2024-role} for critiques of residualization in the context of predictability effects on reading times.}:

\begin{equation}
\textrm{Relative Backward Predictability} = \underbrace{\log p(w_t \mid C_{>t})}_{\textrm{Backward Conditional Probability}} - \underbrace{\log p(w_t \mid C_{<t})}_{\textrm{Forward Conditional Probability}}
\label{eq:relBw}
\end{equation}

Both relative backward predictability and conditional PMI partial out the standalone predictive contribution of the past, but differ in that conditional PMI preserves the joint effects of past and future, whereas relative backward predictability treats past and future as independent sources of predictability. From an information-theoretic perspective, relative backward predictability can be viewed as a log-likelihood ratio that represents how much \emph{more informative} the upcoming sequence is about the current word \emph{compared} to the past context. A positive value of relative backward predictability reflects that the future context outperforms the past in terms of predicting the current word while a negative value reflects the opposite. 

It should be noted that relative backward predictability can be viewed as an \emph{asymmetric} formulation of future context predictability since it reflects an assumption that it is the future context that influences the current word. By contrast, conditional PMI is \emph{symmetric}: it measures the association between the current word and future sequence given past context, but is agnostic about the direction of influence between the current word and the future context. For a full comparison, we can also consider the unconditional PMI in Eq. \ref{eq:uncond-pmi}, which is an alternative symmetric formulation of future context predictability that, like relative backward predictability, assumes independent effects of past and future. We discuss effects of unconditional PMI relative to other formulations of future context predictability in Appendix~\ref{app:study1-model-comps}.

\subsection{Language Modeling} \label{sec:language-modeling}
In addition to the theoretical and modeling considerations addressed above, the estimation of forward predictability, backward predictability, and PMI presents a non-trivial challenge. Much of the prior work on probabilistic reduction focused primarily on bigram contexts, where both forward and backward predictability could be estimated from the joint frequency of the bigram sequence and the respective unigram frequencies of the individual words \citep{pluymaekers2005articulatory,bell2009predictability}. More recently, among studies of speech that have investigated the effects of longer contexts on reduction, it has been common practice to train separate \emph{neural} language models for forward (left-to-right) ($\overrightarrow{p}$) and backward (right-to-left)($\overleftarrow{p}$) predictability, with the backward predictability model being trained on a reversed corpus \citep{dammalapati-etal-2021-effects,harmon2021theory,ranjan2022linguistic}.

\begin{figure*}[t]
  \centering
      \includegraphics[width=0.90\textwidth]{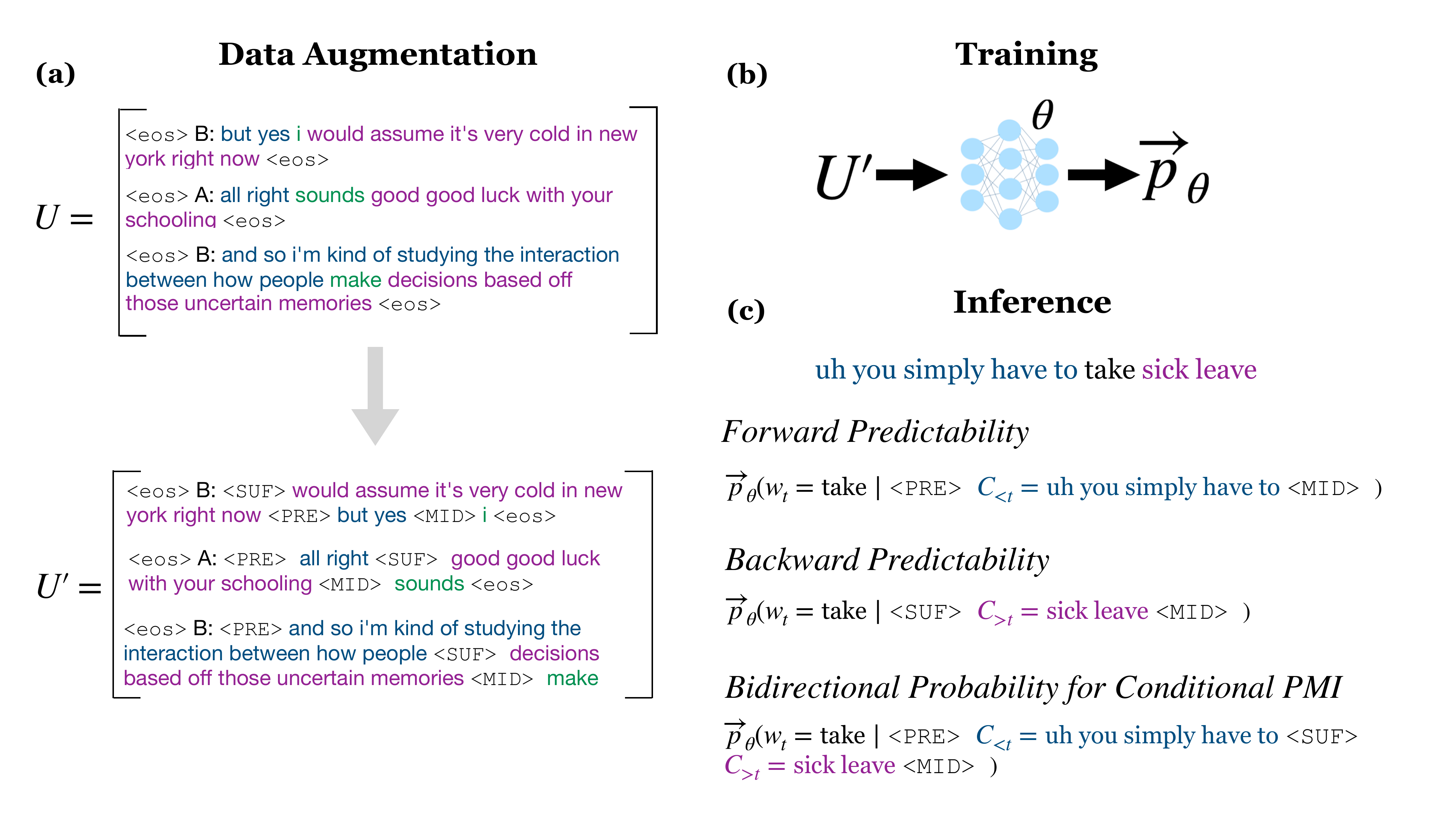}     \caption{Overview of the process for estimating contextual predictability variables from a custom-trained language model (LM). \textbf{(a)} Data augmentation process for enabling estimation of all three probabilities from an autoregressive LM. $U$ is the original corpus of utterances while $U'$ is the augmented corpus. Each utterance ($u$) in $U$ was transformed by uniformly sampling a position in the utterance, selecting the word in that position, and appending this word to the end of the utterance. The past and future context with respect to the original word position are demarcated using $\texttt{<PRE>}$ and $\texttt{<SUF>}$ tokens, and the transposed word is preceded by a $\texttt{<MID>}$ token. In $50\%$ of the utterances, the positions of the $\texttt{<SUF>}$ and $\texttt{<PRE>}$ contexts were swapped. This was done in accordance with prior work, which has found that changing the order of the preceding and following sequences improves estimation of infill probabilities \citep{bavarian2022efficient}. See Appendix~\ref{app: model-training-and-eval} for an algorithmic implementation of this process. \textbf{(b)} $U'$ serves as the training input to a randomly initiated GPT-2 language model parameterized by $\theta$. \textbf{(c)} An illustration of the inference process for estimating forward, backward, and bidirectional probabilities from the trained GPT-2 model $\overrightarrow{p}_{\theta}$.}
\label{fig:language_modeling}
\end{figure*}

However, this practice is not suitable for estimating our measures. First, estimating conditional PMI involves estimating the bidirectional or \emph{infill} probability ($\overleftrightarrow{p}$), which cannot be feasibly estimated from a next-word prediction-based or autoregressive language model.\footnote{Existing pre-trained LMs capable of estimating masked bidirectional probabilities, such as BERT \citep{devlin2018bert}, are not suitable for this task, because they not only differ from autoregressive models in terms of training data, but also in the training objective and model architecture. Furthermore, masked LMs such as BERT are not generative models and do not define a joint distribution over words.} Second, if we measure forward and backward predictability using separately-trained models, we introduce a potential confound, because the probability distribution modeled by a neural LM depends on both the nature of the neural architecture itself and the variability introduced in the training process (for example, the contents of the random mini-batches used in training by stochastic gradient descent, and the random initialization of parameters). If one trains separate forward and backward LMs, their outputs may not only reflect differences in the training input, but also additional sources of variance resulting from the training process \citep{fehlauer2025convergence}.

For these reasons, a principled comparison between the contextual predictability variables requires that forward, backward, and infill probabilities all be derived from a \emph{single} model $\overrightarrow{p}_{\theta}$. To train such a model, we adapt prior work on enabling infill probability estimation from autoregressive LMs \citep{donahue-etal-2020-enabling,bavarian2022efficient}. In principle, this involves selecting a word at random within the utterance, and re-arranging the positions of the word, its preceding context, and following context as follows (see also Figure \ref{fig:language_modeling}):

\begin{enumerate}
\globalitem  Original utterance: So this is the \textbf{first} time I did this conversation 
\globalitem Augmented training input: \texttt{<PRE>} So this is the \texttt{<SUF>} time I did this conversation \texttt{<MID>} \textbf{first} \texttt{<eos>}
\end{enumerate}

Here, the selected word (\textbf{first}) is moved to the end of the utterance such that the model learns to predict this word given both the past and future context that appear \emph{before} it, demarcated using \texttt{<PRE>} and \texttt{<SUF>} tokens, respectively. This allows us to use an autoregressive LM trained on next-word prediction while supplying it with bidirectional context. Furthermore, previous work has shown that training or fine-tuning LMs on naturalistic dialogue improves the model's sensitivity to speech-related behaviors compared to those trained on written text alone \citep{umair2024can}. Since our study focuses on signatures of production difficulty in naturalistic conversations, we train a GPT-2 \texttt{small} (124 million parameter) transformer language model \citep{radford2019language} and a word-level tokenizer on the CANDOR corpus of spontaneous speech \citep{reece2023candor}. Additionally, we also prepend each utterance with a speaker tag (A or B to indicate a change in turn) to account for partner-tracking, which has been shown to affect predictability estimates in conversational speech \citep{warnke2023top}. Model and training-related hyperparameters were evaluated through grid search, and the model that yielded the lowest perplexity on the Switchboard corpus was selected for deriving the probabilistic measures (see Appendix ~\ref{app: model-training-and-eval} for training details). The inference process for estimating forward predictability, backward predictability, and conditional PMI\ from this model ($\overrightarrow{p}_{\theta}$) is detailed in Figure~\ref{fig:language_modeling}(c). Log-transformed forward and backward probabilities were plugged into Eq. ~\ref{eq:relBw} to compute relative backward predictability. Likewise, conditional PMI\ was computed by plugging in log-transformed bidirectional and forward probabilities into Eq.~\ref{eq:PMI}. Further details of the data augmentation and model training process can be found in Appendix \ref{app: model-training-and-eval}. All the contextual predictability measures used in both our studies were estimated using this language modeling approach.

To evaluate how closely forward and backward probabilities obtained from this training approach aligned with those estimated from separate models, we also trained separate forward ($\overrightarrow{p}_{\psi}$) and backward ($\overleftarrow{p}_{\phi}$) GPT-2 models on the unmodified CANDOR corpus. Forward probabilities estimated from $\overrightarrow{p}_{\theta}$ were highly correlated with those obtained from $\overrightarrow{p}_{\psi}$ ($r = 0.85, p < 0.001$), and a similar correlation was observed with backward probabilities obtained from the infill-trained ($\overrightarrow{p}_{\theta}$) and backward-trained ($\overleftarrow{p}_{\phi}$) models ($r = 0.83, p < 0.001$). This suggests that our infill-trained model provides reliable estimates of the predictability values derived from separate models. 

The modeling paradigm described in this section improves upon alternatives involving separate models in that it enables computationally tractable estimation of forward, backward, and infill probabilities from the same underlying distribution $\overrightarrow{p}_{\theta}$. We want to emphasize that we do not consider the infill-augmented GPT-2 as instantiating a particular cognitive claim about how speakers estimate probabilities during on-line production. Rather, we view the various probabilistic quantities estimated from this model as approximating the kinds of contextual information that may be available to speakers during incremental production, which may be arrived at by many different means.

We choose to use the GPT-2 architecture for estimating probabilities because this provides the most accurate estimate of all the information in the context that might be available to a producer. An open question in the literature concerns the amount of context appropriate for explaining phonetic reduction, and recent work by \citet{jacobs2025less} suggests that highly local $n$-gram–based estimates of forward and backward predictability may be sufficient to account for reduction, when combined with information about word position or prosodic structure. 

Though interesting, we primarily leave questions about the explanatory adequacy of different context window sizes
to future work, although we also report analyses with $n$-gram predictors in Appendix~\ref{app: baseline}.

\section{Study 1: Revisiting Predictability Effects on Word Durations} \label{sec:study1}
Previous work on backward predictability effects in production has largely focused on modeling disfluency \citep{bell2003effects,dammalapati-etal-2019-expectation,dammalapati-etal-2021-effects,harmon2021theory} or articulatory realization in speech \citep{bell2009predictability,hashimoto2021probabilistic,ranjan2022linguistic}. Among studies of articulatory reduction, a robust finding is that backward predictability often emerges as a stronger contextual predictor of reduction than forward predictability in both spontaneous and read-aloud speech \citep{bell2009predictability,hashimoto2021probabilistic,ranjan2022linguistic}. A notable result reported by \citet{bell2009predictability} is that bigram backward predictability strongly correlates with reduced durations in content and mid-to-low-frequency words, but does not significantly modulate reduction in high-frequency function words, which are instead highly sensitive to bigram forward predictability. By comparison, \citet{hashimoto2021probabilistic} finds that predictability from the preceding morpheme also affects the articulatory realization of content morphemes in conversational Japanese speech, although backward predictability remains stronger overall. Furthermore, \citet{ranjan2022linguistic} observe that the asymmetric effects of forward and backward predictability across lexical classes reported by Bell et al. are not borne out in the durations of function versus content words in read-aloud Hindi speech.

In this study, we aim to revisit the analysis of \citet{bell2009predictability} with two objectives in mind. First, we seek to conduct a controlled comparison of relative backward predictability and conditional PMI to examine how the assumption of (in)dependence between the past and future context affects the magnitude of the future context predictability effect. Because prior work has revealed inconsistent effects of forward and backward predictability on the realization of function versus content words, a secondary aim of this study is to re-examine the effects of contextual predictability and lexical class on articulatory reduction.

\subsection{Methods}
\subsubsection{Materials} Word durations were extracted from Switchboard NXT annotations \citep{Calhoun2010TheNS,godfrey1992switchboard}, which provides word-level alignments and Penn Treebank part of speech (POS) tags \citep{marcus1993building} along with Tone and Break Indices (ToBI) and disfluency annotations. A key consideration in modeling predictability effects on word duration in naturalistic production concerns the inclusion of disfluencies. Past work has demonstrated that the occurrence of disfluencies correlated with enhanced word duration \citep{dammalapati-etal-2019-expectation}, particularly in the case of function words \citep{foxtree1997pronouncing,bell2003effects}. Consistent with \citealt{bell2009predictability}'s analysis, we exclude words in disfluent contexts by limiting our analysis to only include utterances without repair or reparandum annotations, yielding a total of 435,795 words.

\subsubsection{Statistical Analysis}
\rr{\textbf{Control variables:} We retain the covariates used by \citet{bell2009predictability}, including word length (in syllables), speaker age, and speaker sex. Two of \citeauthor{bell2009predictability}'s controls are prosodic: speech rate and a binary indicator of whether the word lies at the boundary of an intonational phrase (IP). Speech rate, measured in syllables per second, was computed over the entire utterance. Although \citet{bell2009predictability} coded both phrase-initial and phrase-final position separately, phrase-initial position did not significantly improve their control model, and they therefore retained only the phrase-final indicator in their core analysis. This asymmetry is consistent with the broader literature, which has documented robust effects of phrase-final lengthening on segmental and word duration \citep{wightman1992segmental,turk2007multiple,clopper2018exploring}. Because our analysis is intended as a replication of, and comparison against, \citet{bell2009predictability}, we adopt their control variables to preserve interpretability and comparability with their findings. The corpus, however, provides ToBI annotations for only a smaller subset of the data; we therefore exclude the phase-final boundary variable from the analysis reported in this section. See Appendix~\ref{app:replication-prosody} for a replication of this analysis with this additional prosodic control.\footnote{Findings about the explanatory power of relative backward predictability and conditional PMI were qualitatively replicated even after including a binary variable encoding whether the word lies at an IP boundary; however, we note some differences in the effect sizes of the probabilistic predictors, which we discuss in further detail in Appendix~\ref{app:replication-prosody}.}}

\textbf{Probabilistic variables:} Lexical frequency was operationalized as the decontextualized or unigram predictability of a word, which we estimate from a count-based $n$-gram model with Laplace smoothing. To ensure consistency between the language statistics that generate the unigram and contextual predictability variables, the unigram model was also trained on the CANDOR corpus. All of the contextual predictability variables, including forward predictability, relative backward predictability, and conditional PMI, were estimated from the single-model setup described in Section \ref{sec:language-modeling}.\footnote{We also provide replications of Study 1 with bigram predictors in Appendix \ref{app: baseline}. Under the modeling assumptions adopted in our main analysis, we do not observe stronger effects of bigram predictors over predictors estimated from GPT-2. However, as \citet{jacobs2025less} point out, these results may be sensitive to modeling choices such as the source of the unigram estimate (SUBTLEXus versus CANDOR), the choice of neural language model (separately trained forward and backward models versus single infill-trained setup), or other covariates such as word position versus speech rate.}.
 
\textbf{Regression methodology:} We use linear mixed-effects regression models \citep{barr2013random} to fit word duration (in milliseconds), with the maximally converging model below serving as the baseline: \\

Baseline model: Duration  $\sim$  log unigram predictability + log forward predictability + word length + speech rate + speaker age + speaker sex + 1 $\mid$ speaker identity \\

We then generate two variants of this model with future context predictability: one including relative backward predictability (Model 1a) and another including conditional PMI (Model 1b). Since relative backward predictability is a linear combination of forward and backward predictability (see Eq. \ref{eq:relBw}), the explanatory power of this model is exactly equivalent to that of a model with unmodified backward predictability, despite differences in coefficient values \citep{freedman2009statistical}.\footnote{This was confirmed empirically by fitting regression models with backward and relative backward predictability as predictors and comparing their goodness of fit.} Therefore, we do not consider a variant of the baseline model with unmodified backward predictability. However, see Appendix~\ref{app:study1-model-comps} for a complete comparison between models with unmodified backward predictability, relative backward predictability, unconditional PMI, and conditional PMI.

\begin{table}[h]
\centering
\resizebox{\textwidth}{!}{
  \begin{tabular}{@{}l@{\hskip 8pt}rr@{}}
    \toprule
    & \textbf{Model 1a} & \textbf{Model 1b} \\
    & (with relative backward predictability) & (with conditional PMI) \\
    \addlinespace
    \midrule
    (Intercept) & 145.981(1.85)*** & 129.250(1.82)*** \\
    \textbf{Unigram Predictability} & -23.676(0.12)*** & -24.204(0.12)*** \\
    \textbf{Past Context Predictability} & -3.075(0.11)*** & -1.927(0.11)*** \\
    \textbf{Future Context Predictability} & -32.576(0.40)*** & -37.765(0.62)*** \\
    Word Length (in syllables) & 88.569(0.38)*** & 88.273(0.38)*** \\
    Speech Rate & -32.767(0.14)*** & -33.163(0.14)*** \\
    Speaker Age & 0.070(0.04) & 0.090(0.104) \\
    Speaker Sex:M & -12.035(0.86)*** & -11.117(0.84)*** \\
    \bottomrule
  \end{tabular}
} 
\caption{Regression coefficients from models with relative backward predictability and
conditional PMI as operationalizations of future context predictability. Probabilistic
predictors are bolded. Parentheses denote standard error. $p < 0.001$ (***), $p < 0.01$ (**),
$p < 0.05$ (*), $p > 0.05$ (\emph{ns}).}
\label{tab:DurModelCoeffs}
\end{table}

We then compare these model variants with the goal of identifying the distinct explanatory contribution of future-context predictability and further characterizing the contributions of relative backward predictability and conditional PMI. First, we conduct pairwise model comparisons between the baseline model and Models 1a and 1b to quantify the additional variance explained by including any measure of future context predictability, and to assess which of the two variants yields a greater improvement in explanatory power. We further compare models 1a and 1b to a model with \textbf{both} relative backward predictability and conditional PMI (Model 1c) to examine the extent to which these measures make overlapping versus non-redundant contributions to explaining variance in word durations. Since these regression models are nested, and because the goal of this analysis is to quantify the explanatory power of the two operationalizations of future-context predictability rather than evaluate model complexity, we use the Likelihood Ratio Test to conduct model comparisons. \footnote{We found no difference in the qualitative conclusions obtained from likelihood ratio tests as opposed to a complexity-penalizing model selection criterion such as the Bayesian Information Criterion (BIC). See Appendix ~\ref{app: bic-model-comp} for BIC-based model comparisons.}

Finally, to examine the modulatory effects of lexical class on word duration, we set up additional variants of Models 1a–1b that include interactions between lexical class and the probabilistic predictors. Lexical class was coded as a binary factor with \textsc{FUNCTION} as the reference level. \rr{The class of each word was determined based on the Penn Treebank POS tags provided in the corpus. Nouns, verbs, adjectives, adverbs, and cardinal numbers were categorized as content words while the rest of the tags (excluding disfluencies and symbols) were categorized as function words \footnote{\rr{In certain cases, the same word received different POS tags in the corpus based on context e.g., the word `really' was tagged as RB (adverb) in 166 instances and UH (disfluency) in 46 instances. In this case, only the occurrences of `really' that were tagged as `RB' were mapped to the content class.}}}. This yielded a total of 416,831 words, with 214,990 content words and 201,841 function words.

See Appendix \ref{app:study1-stat-model-specs} for detailed specifications of all the models used in this analysis.

\subsection{Results} \label{sec:study1-results}

\begin{figure*}[t]
    \centering
    \includegraphics[width=\textwidth]{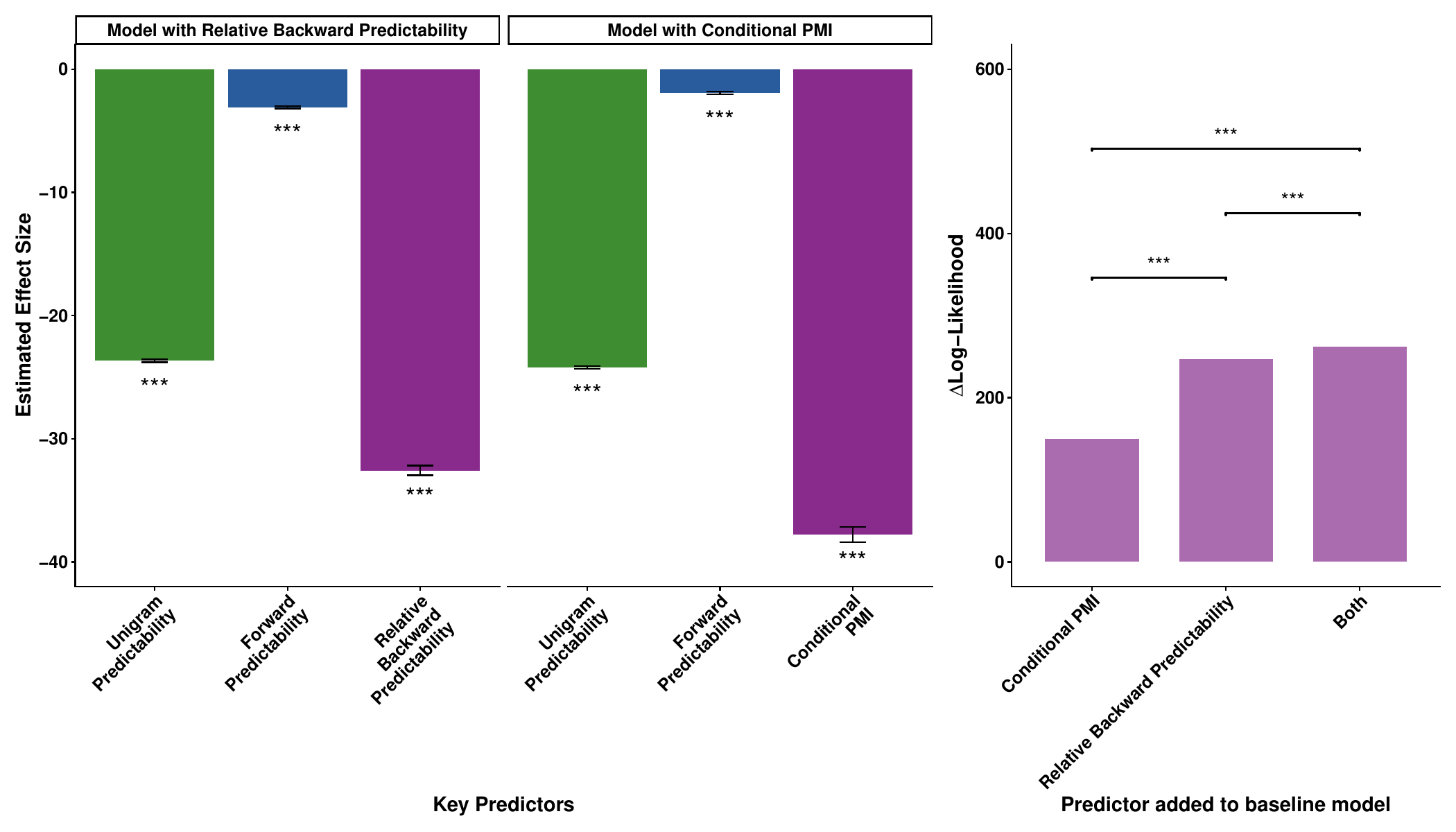}
    \caption{\textbf{(a)} Estimated effect sizes for all probabilistic predictors from models with relative backward predictability and conditional PMI as formulations of future context predictability. Relative backward predictability assumes independence between past and future, whereas conditional PMI assumes conditional dependence. Error bars denote standard error. $p < 0.001$ (***), $p < 0.01$ (**), $p < 0.05$ (*), $p > 0.05$ (\emph{ns}); \textbf{(b)} Delta log-likelihood values obtained from adding future context predictability measures to the baseline model incrementally. Higher values of $\Delta$\textbf{Log-Likelihood} indicate a better fit to the data.} \label{fig:study1-results}
\end{figure*}

\begin{figure*}[t]
  \centering
      \includegraphics[width=\textwidth]{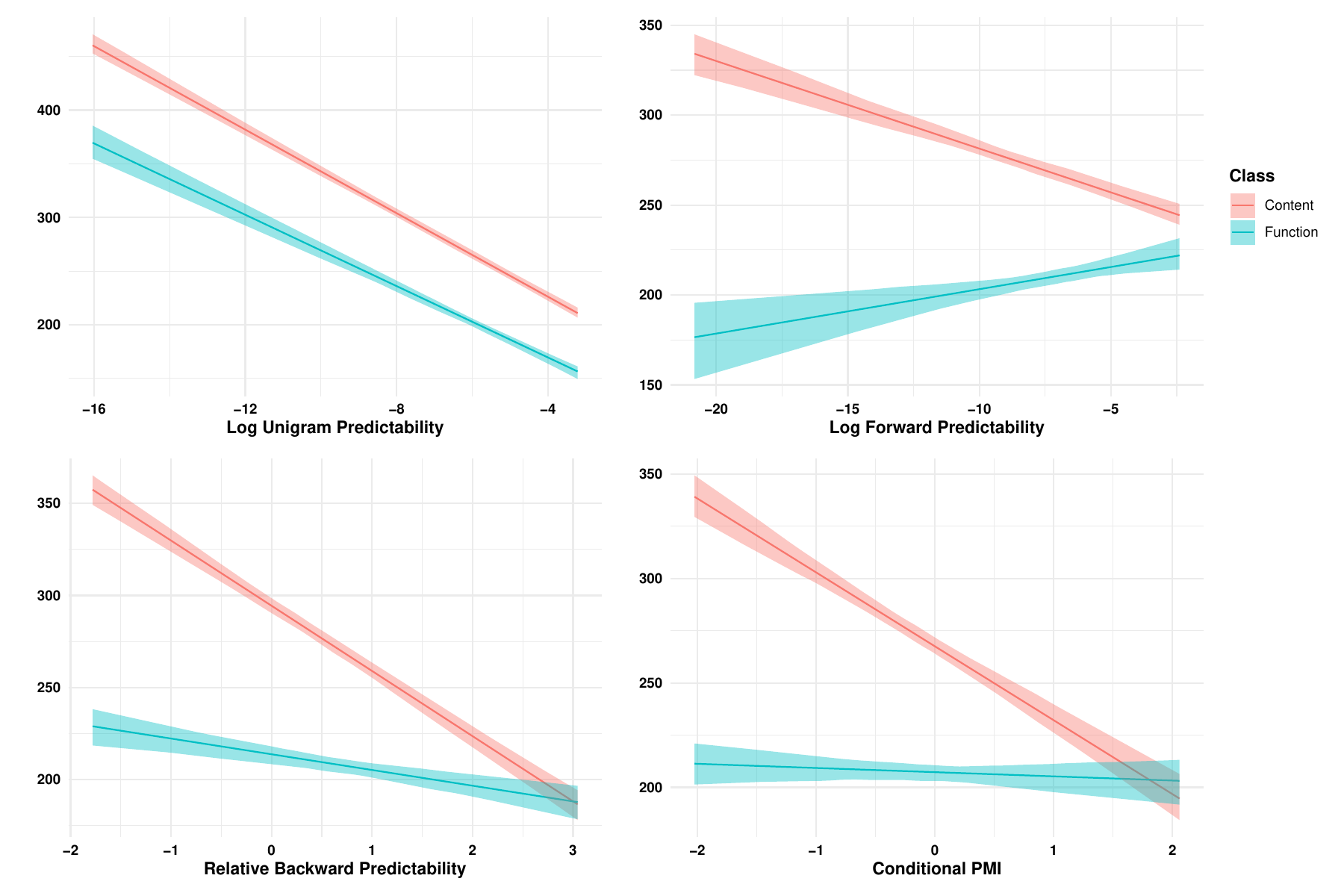}
          \caption{The modulatory effect of lexical category (function/content) on probabilistic reduction, \rr{when controlling for other predictors}. All predictors are log-transformed. 95\% confidence intervals estimated via bootstrapping (N = 1000 simulations). All interactions emerged significant.}
\label{fig:Study1_LexCatProbReduction}
\end{figure*}

\paragraph{Model Comparison:}
Regression coefficients from Models 1a and 1b are summarized in Table \ref{tab:DurModelCoeffs}. Positive coefficients predict lengthening of word duration, whereas negative coefficients reflect reduction. \rr{In both Models 1a and 1b, word length is positively correlated with duration while faster speech rate is associated with reduction}. 

As expected, all three probabilistic variables exhibit a significant inverse relationship with word duration (Figure~\ref{fig:study1-results}a). First, the frequency effect was consistently replicated: higher unigram predictability correlates with reduced word duration in both model variants. Predictability from the past context is likewise negatively correlated with duration, although this effect is smaller than that of both unigram predictability and predictability from the future. Among the three probabilistic variables, predictability from the future emerged as the strongest predictor of reduction, with both relative backward ($\beta =  -32.576, \ SE = 0.396, p < 0.001$) and conditional PMI ($\beta = -37.765, \ SE = 0.618, \ p < 0.001$) exhibiting an inverse effect on duration.

Results of the incremental likelihood ratio tests are presented in Figure~\ref{fig:study1-results}b. Including predictability from the future improved model fit relative to baseline, both for relative backward predictability (Model 1a: $\deltaLoglik = 3363.199, \ \chi^{2} = 6726.5,\  p < 0.001$) and conditional PMI (Model 1b: $\deltaLoglik = 1856.581, \ \chi^{2} = 3714, p < 0.001$). Moreover, a comparison between models 1a and 1b reveals that relative backward predictability explains more variance in word duration than conditional PMI ($\deltaLoglik = 1506.618, \  \chi^{2} = 3012.5, \ p < 0.001$). However, the model with both conditional PMI and relative backward predictability yielded a small albeit significant increase in explanatory power compared to the model with relative backward predictability ($\deltaLoglik = 122.166, \  \chi^{2} = 245.6, \ p < 0.001$), hence emerging as the strongest performing model of the three.

\paragraph{Effects of lexical class and predictability on word duration:} 
All regression coefficients from these analyses are reported in Appendix \ref{app:lexClass-analysis} and interactions between lexical class and predictability are visualized in Figure \ref{fig:Study1_LexCatProbReduction} . Across both Model 1a and 1b variants, there is a significant interaction between unigram predictability and lexical class, with unigram predictability exhibiting a stronger effect on reduction for content words compared to function words. A similar pattern holds for forward predictability, where content word durations show greater attenuation when they are predictable from the past. For function words, the simple effect of forward predictability is small and positive, indicating attenuation and even possible reversal of the reduction effect once other sources of predictability are controlled for. Finally, the regressions reveal significant interactions between lexical class and future-context predictability, both for relative backward predictability and conditional PMI. Notably, similar to unigram predictability and predictability from the past, content word durations are more strongly modulated by predictability from the future than function word durations.

\subsection{Discussion}
The key findings of this study can be summarized as follows. First, relative backward predictability and conditional PMI yield qualitatively similar effects on reduction, with both measures exhibiting an inverse relationship with word duration. Crucially, predictability from the future, regardless of operationalization, emerges as the strongest probabilistic correlate of reduced durations in our analysis, followed by unigram predictability. By comparison, predictability from the past shows a small albeit robust effect across all model variants.

However, the two operationalizations of future context predictability diverge quantitatively. Our model comparisons reveal that relative backward predictability provides a better fit to word durations than conditional PMI. This pattern is further supported by the observation that, in the model including both measures, relative backward predictability emerges as the strongest contextual predictor of word duration, whereas conditional PMI shows a smaller effect than both relative backward and unigram predictability. It should be noted, however, that backward predictability does not subsume conditional PMI in explaining variance in word duration, as including conditional PMI still yields a significant gain in explanatory power. This suggests that the two operationalizations of future-context predictability are non-redundant variables: conditional PMI contributes complementary information above and beyond what is captured by relative backward predictability. In other words, word duration is influenced to some extent by the predictability of the word given both the past and the future \emph{jointly}, beyond what could be explained by predictability from the past and future separately. We revisit the discussion about these two measures in Sections \ref{sec:study2-discussion} and \ref{sec:general-discussion}. 

A second key finding of this study concerns the function–content asymmetry in wordform variation. Notably, \citealp{bell2009predictability} reported that the effects of forward predictability on articulatory duration were limited to high-frequency function words,\footnote{These include ten of the most high frequency function words i.e., \emph{a},\ \emph{the}, \emph{in}, \emph{of}, \emph{to}, \emph{and}, \emph{I}, \emph{it}, \emph{you}.} with no significant effect on the duration of mid-to-low frequency function and content words. By comparison, backward predictability emerged as the strongest contextual predictor of content and mid-to-low frequency function words. However, our results did not replicate these asymmetric effects of past and future context predictability across lexical classes. Instead, we observe that predictability from both directions predicted reduction in content word durations. The effect of lexical class was, however, borne out in the sensitivity to predictability-induced reduction: content words exhibit stronger predictability-induced reduction across all sources of predictability (unigram, past, and future context), whereas effects for function words are substantially attenuated. Therefore, while both function and content words are sensitive to probabilistic structure regardless of whether predictability is conditioned on previously produced context or unrealized future representations, the degree of sensitivity appears to be modulated by lexical class.

In function words, unigram predictability emerges as the strongest probabilistic correlate of word duration, suggesting that reduction in function words may primarily reflect context-independent lexical accessibility or routinization \citep{foxtree1997pronouncing,bell2003effects,bybee2002word,kapatsinski2010frequency}. The effects of contextual predictability on the duration of function words is more nuanced. First, we obtain a mild positive relationship between forward predictability and duration of function words when interactions between lexical class, unigram predictability, and future context predictability are included in the model. However, the positive coefficient for forward predictability in function words should be interpreted with caution, as a simpler model without interactions yields the expected negative relationship, suggesting this is likely an artifact of including multiple correlated interaction terms in the model. Comparing the two operationalizations of future context predictability, we find that reduction in function word durations is most strongly driven by relative backward predictability while sensitivity to conditional PMI is attenuated. We discuss the conceptual implications of these results for the two operationalizations of future context predictability in Section \ref{sec:general-discussion}.

It bears mentioning that our analysis of the function–content distinction in probabilistic reduction differs from \citet{bell2009predictability} along two dimensions: the size of the context window and the language models used for estimating predictability. First, \citet{bell2009predictability} restricted contextual windows to the immediately preceding or following word and estimated forward and backward predictability using count-based bigram models. In comparison, our study employed the GPT-2 language model, which conditions on broader context and produces smoothed predictability estimates derived from a continuous representational space \citep{kapatsinski2026transformers}. Other recent studies using neural language models (e.g., LSTMs or transformers) have likewise reported significant effects of forward predictability or surprisal on content word durations \citep{ranjan2022linguistic,clark2025relationship}. One possible explanation for this divergence is that these results may be sensitive to the differences in predictability estimates derived from discrete $n$-gram vs continuous neural language models.

\section{Study 2: Modeling Substitution Errors in Naturalistic Productions} \label{sec:study2}
The above study presents a controlled comparison between two alternative operationalizations of future context predictability, both of which exhibit qualitatively similar behavior to backward predictability in modulating articulatory duration. While we did not observe differential effects of past and future context predictability in the reduction of function versus content words, word durations were nonetheless more sensitive to predictability from the future than from the past. This observation is consistent with broader findings that backward predictability emerges as the strongest contextual predictor of other planning difficulties such as filled pauses and repetitions \citep{shriberg1996disfluencies,dammalapati-etal-2019-expectation,dammalapati-etal-2021-effects,harmon2021theory}. 

However, modeling reduction or disfluency may not be the most suitable paradigm for investigating the distinct influences of past and future context on lexical planning. These phenomena primarily index the ease of cognitive processing during production, with unpredictability from either direction correlating with an increase in planning difficulty. Therefore, they offer a limited window into the sentence planning mechanisms by which context influences speaker choices during lexical planning. For example, consider the following utterances from the Switchboard corpus:

\begin{enumerate}
    \globalitem Well what happens is that if people start $\colorbox{red!30}{becoming}$ $\colorbox{green!30}{having}$ chronic illness and and things like that \label{ex:subs1}
    \globalitem Uh you simply have to $\colorbox{red!30}{take}$ $\colorbox{green!30}{accumulate}$ your sick leave. And take your sick leave \label{ex:subs2}
\end{enumerate}

\noindent Ex. (\ref{ex:subs1}) and (\ref{ex:subs2}) illustrate the distinct pressures imposed by past and future context on word choice. Whereas the preceding context may steer the speaker into selecting \emph{becoming} in (\ref{ex:subs1}), this choice is incompatible with the upcoming material, which strongly favors \emph{having}. Similarly, \emph{take} is a more frequent and predictable choice compared to \emph{accumulate} in (\ref{ex:subs2}), despite being incompatible with both the future context and the speaker's message. In both cases, incompatibility with future plans and intended semantics incentivizes the speaker to initiate a repair. 

Prior work has shown that lexical accessibility and semantic alignment can similarly exert competing influences on word choice, leading to mis-selections or \emph{good-enough choices} \citep{rapp2002reason,ferreira2003phonological,koranda2022good,goldberg2022good}. Because speakers operate under information processing constraints, lexical selection may reflect a trade-off between automatic or \textbf{goal-invariant} strands of processing that prioritize \emph{economy of effort} and controlled or \textbf{goal-directed} processing that aims to satisfy the speaker's communicative intent \citep{bock1982toward,ferreira2002central,hartsuiker2017automaticity,futrell2023information}. Through independent manipulation of availability and alignment, \citet{koranda2022good} demonstrated empirical evidence of this trade-off in a gamified word production paradigm: speakers often preferred highly frequent but semantically imprecise alternatives over infrequent yet semantically precise choices to describe a production target.

Beyond word production, previous work has shown that highly incremental (i.e., left-to-right) planning is heavily influenced by conceptual and lexical accessibility \citep{bock1980syntactic,mcdonald1993word,griffin2001gaze,gleitman2007give,dell2008saying,iwasaki2011incremental,Momma2019BeyondLO}. This preference for efficiency is reflected in the \textsc{principle of immediate mention} \citep{ferreira2000effect}: speakers choose to prioritize the overt production of highly available lexical representations to routinely free-up working memory resources \citep{ferreira2000effect,ferreira2002incremental,slevc2011saying,christiansen2016now}. Under this accessibility-driven strategy, the previously produced context may steer the speaker toward highly predictable lexical choices, potentially at the cost of alignment with desired semantics or upcoming context.  In contrast, advance planning of the upcoming sequence may impose greater cognitive demands on the speaker since lexical representations need to be planned in advance and sustained in working memory for much longer \citep{wagner2010flexibility,lee2013ways,Momma2019BeyondLO}. 

Therefore, one possible interpretation that we seek to explore in this study is that past and future context exert competing influences of word choice, reflecting aspects of goal-invariant and goal-directed aspects of processing, respectively. Previously, the study of trade-offs in word choice has been restricted to single-word production since it enables precise manipulation of availability and alignment, but neglects the role of context. In this study, we focus on modeling word choice in naturalistic substitution contexts such as (\ref{ex:subs1})-(\ref{ex:subs2}), where the observed self-repair (e.g., $\colorbox{green!30}{having}$ in Ex. \ref{ex:subs1}) provides a reasonable approximation of the speaker's intent or production target. While previous studies have modeled the effects of frequency on the choice of substitution errors \citep{kapatsinski2010frequency} and the effects of contextual predictability on the \emph{occurrence} of substitution disfluencies \citep{dammalapati-etal-2019-expectation}, our proposed paradigm aims to predict the \emph{content} or identity of the substitution error using a set of theoretically-motivated predictors that index 
lexical accessibility, semantic and form-based alignment with the intended word, and context-based mechanisms of sentence production.

\subsection{Methods}
\subsubsection{Materials}
Utterances with naturally-occurring lexical substitutions and self-repairs were identified and extracted from Switchboard NXT annotations using three criteria. First, utterances with an unequal number of \emph{reparandum} and \emph{repair} words were excluded to avoid instances where the speaker may have revised the structural plan of the sentence. Second, we consider two types of repair: (i) where the self-repair immediately followed the reparandum (Ex. \ref{ex:subs1}) and (ii) where the self-repair was preceded by a repetition disfluency or filled pause (see Ex. \ref{ex:subs3} below).

\begin{enumerate}
\globalitem So until I see the entire quote old guard of the soviet $\colorbox{red!30}{military}$ of the soviet $\colorbox{green!30}{government}$ completely roll over and disappear preferably buried, I still consider them a threat
\label{ex:subs3}
\end{enumerate}

Finally, we apply a grammatical category constraint, which restricts our set of utterances to those where the reparandum and repair had the same Penn Treebank part of speech category. The selected utterances were then processed into utterance frames consisting of the context preceding the error ($C_{<t}$) and the context following the error ($C_{>t}$) with intervening disfluencies removed. Utterances that contained $N$ substitution errors were processed into $N$ distinct frames, with each frame containing one substitution position (see Appendix \ref{app:subs-examples} for an example). This process yielded a total of 773 substitution utterances.

\subsubsection{Model of Substitution Errors} \label{sec:study2-model}
We propose a framework for modeling substitution errors that captures the distinct influences of lexical frequency, contextual predictability, and communicative utility on lexical mis-selection. Within this framework, we aim to predict which word is produced at the substitution position $t$ in the utterance context, framing this task as a logistic regression model that treats the observed error as the positive class and all other alternatives as negative instances. In practice, each word in the vocabulary is treated as a candidate for production at the substitution position, and the model estimates which properties of candidate words increase the probability that a given word is produced as the substitution error. Therefore, predictors in this model indicate whether a given property facilitates or inhibits lexical mis-selection. 

In addition to lexical frequency and contextual predictability, we also consider the \textbf{communicative utility or reward} of choosing a given word, defined as its proximity to the self-repair or intended target $w^{*}_{t}$ along both semantic and phonological dimensions. Therefore, variables in this model are functions of a possible word $w_t$, the entire utterance context $C$, and the intended target $w^{*}_{t}$. 

\paragraph{Operationalizing communicative reward:} Although the observed self-repair provides an approximation of the speaker's communicative intent, several lexical representations may be compatible with the intended semantics. Because semantic and phonological processes interact and jointly constrain lexical selection within the production system, speakers may trade off semantic alignment for lexical accessibility, particularly when multiple representations approximate the intended lexical semantics \citep{dell1992stages,cutting1999semantic,rapp2000discreteness,ferreira2003phonological,goldrick2011interaction,koranda2022good}. Even when speakers have access to the precise semantic and syntactic properties of a lexical representation, \emph{Tip-of-the-Tongue (ToT)} phenomena demonstrate that they may nevertheless experience difficulty accessing its phonological form \citep{brown1966tip,caramazza1997many}. Consequently, speakers may experience uncertainty about both the intended semantics and the phonological form during online production.

To simulate this uncertainty, we generate a ``noisy'' representation of the production target by introducing random perturbations to the semantic and phonetic representation of the self-repair ($w^{*}_{t}$). We then measure the communicative utility of alternative lexical representations against this noisy version of the target ($\hat{w}_{t}$).
Conceptually, this allows us to model uncertainty about the speaker's intended lexical target rather than assuming that the observed self-repair corresponds exactly to that target. Methodologically, it also enables us to include the repair as an alternative in the model, since the distance between the observed repair and the noisy target representation will generally be non-zero by construction.

Semantic representations are estimated by mapping words to real-valued word embedding vectors. We derive these vectors by fitting \texttt{fasttext} embeddings \citep{bojanowski2016enriching} to the CANDOR corpus, which yield 100-dimensional semantic vector representations. To generate a noisy semantic representation of the target, Gaussian noise is injected into the target word vector as follows:
\begin{equation}
    \hat{\textbf{w}}_{t} = \textbf{w}^{*}_t + \boldsymbol{\epsilon},
\end{equation}
where $\textbf{w}^{*}_t \in \mathbb{R}^{100}$ is the vectorized representation of the target and $\boldsymbol{\epsilon} \sim \mathcal{N}(0,0.1) \in \mathbb{R}^{100}$ is the noise vector. We then operationalize the semantic distance as the cosine distance between the noisy target vector $\hat{\textbf{w}}_{t}$ and a given word $\textbf{w}_i$.

\begin{figure*}[t]
  \centering
      \includegraphics[width=\textwidth]{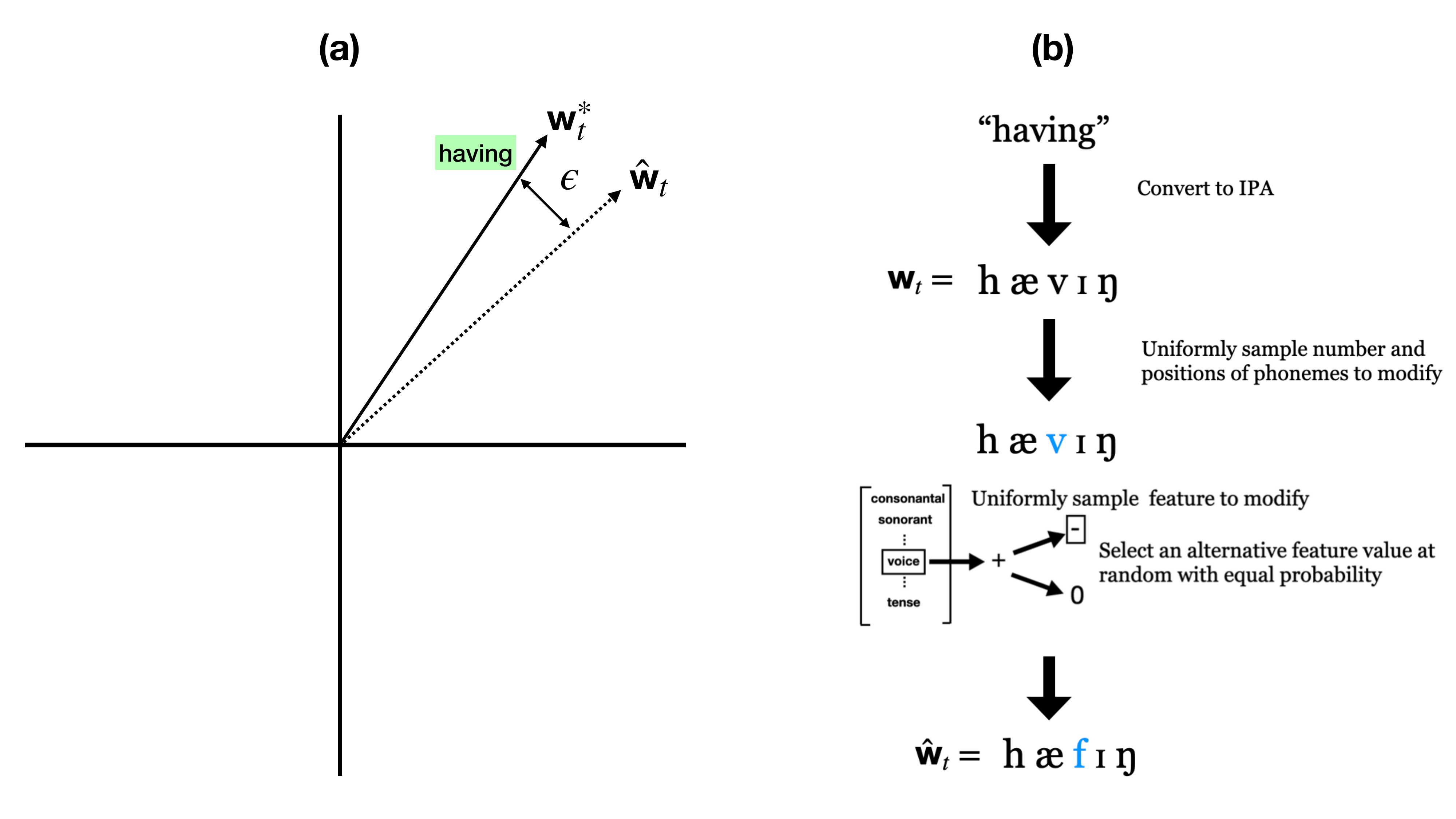}
          \caption{An illustration of the process for computing: \textbf{(a)} the noisy semantic target ($\hat{\textbf{w}}_{t}$) by injecting noise ($\boldsymbol{\epsilon}$) into the target word's embedding vector ($\textbf{w}^{*}_{t}$) and \textbf{(b)} the noisy phonetic representation of the target using a procedure for injecting noise into the phonetic feature-based target representation. See also Appendix \ref{app:algorithms} for a detailed implementation of (b).}
\label{fig:noisy-distance}
\end{figure*}

We generate a ``noisy'' phonetic representation of the target by uniformly sampling phonemes from the IPA form of the target word, and for each selected phoneme, uniformly sampling a phonetic feature to modify. For each chosen feature, we then randomly sample an alternative categorical value to add uncertainty to the target phonetic representation (see Appendix \ref{app:algorithms} for an algorithmic implementation of this process). Finally, phonetic distance was operationalized as the feature-wise distance between the phonetic representation of a given word $w_i$ and the noisy phonetic representation of the target $\hat{w}_{t}$. We use the \texttt{panphon}  package in Python \citep{Mortensen-et-al:2016} to generate categorical phonetic feature representations of words and to compute the feature-based edit distance between the phonetic representation of a given word and the noisy target. See Figure \ref{fig:noisy-distance} for an example of how these processes generate noisy semantic and phonetic representations of the production target.

The noise process detailed in Figure \ref{fig:noisy-distance}b may yield phonetic representations of the target that do not reflect phonemic, phonotactic, or grammatical category constraints observed in experimentally elicited or naturalistic form-related errors \citep{fay1977malapropisms,baars1975output,goldrick2010mrs}. However, we treat this process as analogous to injecting Gaussian noise into the semantic embedding of the target, which may produce vectors that do not correspond to attested lexical items. As with the noisy representation of the target semantics, our goal is simply to introduce random perturbations into the phonetic form to approximate uncertainty about the intended production target.

\paragraph{Regression methodology} For every utterance, the observed substitution error (e.g., $\colorbox{red!30}{becoming}$ in Ex. \ref{ex:subs1}) is treated as an instance of the positive class whereas all the other words in the vocabulary ($N = 14116$, including the self-repair, e.g., $\colorbox{green!30}{having}$ in Ex. \ref{ex:subs1}) constitute the negative class. We use the \texttt{lme4} package to fit the generalized linear model below: \\

\emph{Produced}$(w_i)\sim$  Log Unigram Predictability + Log Forward predictability + (Noisy) Semantic Distance + (Noisy) Phonetic Distance \label{eq:study2-baseline} \\

where $w_i$ is a given word in the vocabulary, \emph{Produced}$(w_i) = 0$ if the word is not the observed substitution error (i.e., if the word is an alternative word from the vocabulary or the self-repair itself), and \emph{Produced}$(w_i) = 1$ if $w_i$ is the observed error. Put differently, the model learns the weights that best predict the observed lexical mis-selection in context. A positive model coefficient indicates that an increase in the predictor leads to an increase in the log-odds of a word $w_i$ being the observed substitution. That is, it facilitates erroneous production of the word. By comparison, a negative coefficient indicates that an increase in the value of the predictor reduces the log-odds of a word being the observed error, hence inhibiting its production \emph{as an error}. Similar to Study 1, we generate three variants of this model with (i) relative backward predictability (Model 2a), (ii) conditional PMI (Model 2b), and (iii) both relative backward predictability and conditional PMI (Model 2c) to examine the unique and redundant contributions of these measures. See Appendix \ref{app:study2-stat-model-specs} for specifications of all the logistic regression models used in these analyses.

\subsection{Results} \label{sec:study2-results}
\begin{figure*}[t!]
    \centering
    \includegraphics[width=\textwidth]{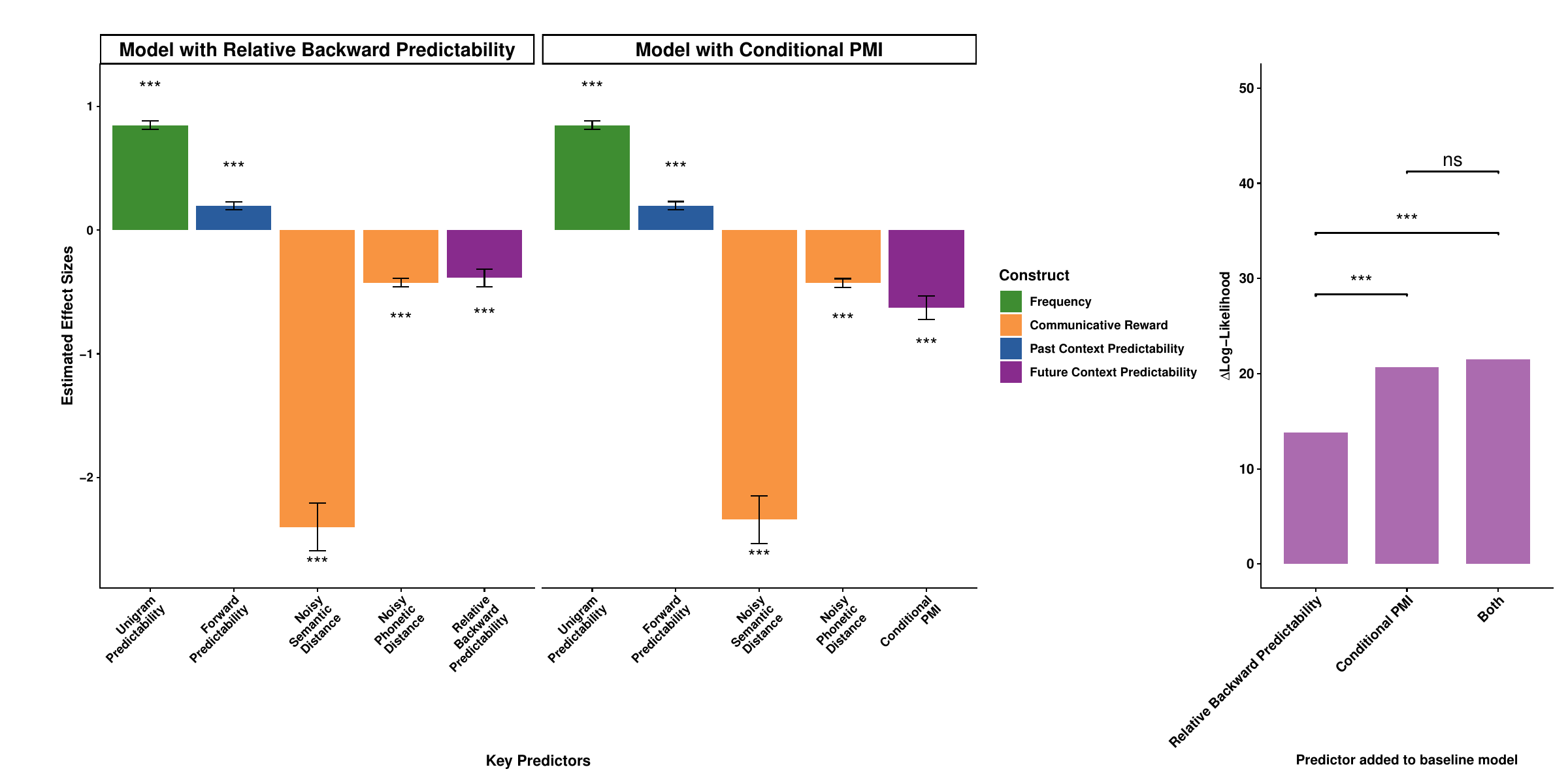}
    \caption{\textbf{(a)} Log-odds coefficients from logistic regression models with relative backward predictability and conditional PMI as operationalizations of future context predictability. Relative backward predictability assumes independence between past and future, whereas conditional PMI assumes conditional dependence. Error bars denote standard error. Positive coefficients indicate that an increase in the predictor facilitates mis-selection of a given word, while negative coefficients inhibit its production as an error. $p < 0.001$ (***), $p < 0.01$ (**), $p < 0.05$ (*), $p > 0.05$ (\emph{ns}); \textbf{(b)} Delta log-likelihood values obtained from adding future context predictability measures to the baseline model incrementally. Higher values of $\Delta$\textbf{Log-Likelihood} indicate a better fit to the data.} \label{fig:study2-results}
\end{figure*}

\paragraph{Factors affecting substitution choices:} All log-odds coefficients from Models 2a and 2b are presented in Appendix \ref{app:study2-result-tables} and all predictors emerged significant. Here we visualize the effects of frequency, predictability, and communicative reward in Figure \ref{fig:study2-results}a. In both model variants, unigram predictability was associated with increased log-odds of a word being selected as the observed substitution, with forward predictability showing a similarly consistent positive effect. In contrast, both semantic and phonetic distances exhibited inverse effects: greater distance from the target representation reduced the likelihood of a word being produced as an error. Intriguingly, unlike unigram and forward predictability, relative backward predictability and conditional PMI exhibited negative effects, aligning qualitatively with the distance-based measures. Increased predictability from the future was associated with decreased log-odds of a word being selected as the substitution, both for relative backward predictability and conditional PMI.

\paragraph{Model Comparison:} Similar to study 1, we also conduct incremental model comparisons with log-likelihood as a measure of explanatory power. Adding relative backward predictability to the baseline model with unigram predictability, forward predictability, semantic, and phonetic distances led to an improvement in model fit, as evinced by an increase in the log-likelihood ($\deltaLoglik = 13.812, \  \chi^{2} = 27.624, \  p < 0.001$). Similarly, adding conditional PMI to the baseline model also improved its goodness of fit ($\deltaLoglik = 20.656, \ \chi^{2} = 41.312, \ p < 0.001$). A comparison between these two models revealed that the model with conditional PMI emerged a better model of substitution choices than the model with relative backward predictability ($\deltaLoglik = 6.843, \ \chi^{2} = 13.688, \ p < 0.001$). Contrary to the results of Study~1, a comparison between the better model (i.e., model with conditional PMI) and the model with both relative backward predictability and conditional PMI revealed no significant improvements in the substitution model's explanatory power ($\deltaLoglik = 0.845, \ \chi^{2} = 1.689, \ p > 0.05$).

\subsection{Discussion} \label{sec:study2-discussion}
\rr{The aim of the above analysis was to explicate the distinct contributions of lexical, semantic, and contextual factors in word choice in naturalistic substitution contexts. Across utterances, we observe a strong frequency effect consistent with prior findings: higher-frequency words were more likely to surface as errors than the lower-frequency targets that speakers subsequently produced as self-repairs \citep{dell1990effects,shriberg1996disfluencies,kittredge2008effect,kapatsinski2010frequency}. Beyond frequency, intrusions were jointly shaped by lexical accessibility and proximity to the target: words that surfaced as errors tended to be frequent and predictable from the past context, yet also lay closer to the target along both semantic and phonetic dimensions. This tension between semantic alignment and form accessibility has been demonstrated in prior research \citep{ferreira2003phonological,koranda2022good} in highly-controlled single-word production paradigms. Our analysis reproduces this pattern at scale, in naturalistic speech, using distributional and phonetic feature-based estimates of 
semantic and form-based alignment, respectively.}

\rr{Prior work offers two complementary insights into how contextual predictability shapes production errors. First, priming can boost the accessibility of competing alternatives, increasing the likelihood that they are mis-selected \citep{bock1986meaning,rapp2002reason,ferreira2003phonological}. Second, less predictable contexts elicit more disfluencies and are associated with higher error rates \citep{foxtree1995effects,bell2003effects,dammalapati-etal-2019-expectation}. Both of these strands of research concern the \emph{occurrence} of errors. While context is widely assumed to shape lexical selection, to our knowledge this is the first empirical demonstration that both past- and future-context predictability predict the \emph{identity} of naturalistic substitution errors. Using contextual probabilities derived from neural language models, we show that these effects hold once frequency, semantic alignment, and phonetic alignment are controlled for.}

\rr{Notably, we find that past and future context appear to exert opposing influences on lexical mis-selection. Like frequency, higher past context predictability increases the likelihood that a word surfaces as an error, suggesting that highly predictable words may be more prone to erroneous production. Future context predictability, however, decreases this likelihood, suggesting that words more compatible with the speaker's future context are less likely to surface as errors.}

\rr{To illustrate, consider the substitution error in Ex.~\ref{ex:subs3}: \textit{So until I see the entire old guard of the Soviet [military/government] completely roll over and disappear…a threat}. Although the past context is compatible with both the error (\textit{military}) and the target (\textit{government}), the future context is substantially more informative about the target (relative backward predictability(\textit{government}) = 1.30, conditional PMI(\textit{government}) = 0.86) than the error (relative backward predictability(\textit{military}) = 0.65, conditional PMI(\textit{military}) = 0.25). More generally, words that share greater informativity with the future context are less likely to surface as errors—an inhibitory effect qualitatively distinct from the facilitative effects of frequency and past predictability on mis-selection. We discuss interpretations of this effect in Section~\ref{sec:general-discussion}}.

\rr{Both relative backward predictability and conditional PMI emerged as significant predictors of word choice, with larger effect sizes than forward predictability, albeit operating in the opposite direction—inhibiting rather than facilitating error production. Unlike Study~1, the substitution model with conditional PMI outperforms the model with relative backward predictability, and including both predictors yields no improvement in fit, suggesting that relative backward predictability is redundant once conditional PMI is accounted for. A potential explanation is that lexical planning recruits the richer contextual representation captured by conditional PMI, which quantifies how informative the future context is about the present word under the constraints of the observed past. Consequently, relative backward predictability—which ignores the specific past context in which both the future context and the current word are planned—may add little explanatory value once the more contextualized conditional PMI is included.}

\section{General Discussion} \label{sec:general-discussion}
Online production is highly, but not strictly, incremental \citep{levelt1989speaking,ferreira1996better,ferreira2000effect,ferreira2002incremental,wheeldon2006language,wagner2010flexibility,slevc2011saying}: the order in which constituents are planned need not reflect the order in which they are produced \citep{Bock1994LanguageP,ferreira2013syntax,ferreira2007grammatical,lee2013ways,Momma2019BeyondLO}. Consequently, speaker choices are shaped by predictability from both previously produced past context and planned future context. Prior work has operationalized the latter using backward predictability---the conditional probability of the current word given some future context. However, much of our current understanding of this backward predictability effect comes from corpus studies involving $N$-gram and LSTM models \citep{bell2003effects,bell2009predictability,dammalapati-etal-2021-effects,harmon2021theory,ranjan2022linguistic}, which condition on limited context and can produce coarser probability estimates.

Intriguingly, many of these studies have noted stronger effects of backward predictability on symptoms of production difficulty, such as prolonged pronunciation and disfluency. However, the interpretation of backward predictability both as a probabilistic variable and as a cognitive construct has been complicated by a few factors. First, backward predictability is often highly correlated with forward predictability, which makes it difficult to disentengle their distinct contributions on  speaker choices. More importantly, backward predictability captures the predictive effects of future context on the present word, but ignores the context in which both of them occur. Therefore, it can be viewed as a \emph{decontextualized} estimate of how future representations shape the speaker's current choices, since it abstracts away from the particular past context in which those choices are situated.

This work presents an examination of contextual predictability effects, both from the past and future, on two aspects of speaker choice: phonetic encoding and word choice. In addition to clarifying the distinct contributions of past and future context on how speakers choose and encode words, our work addresses the aforementioned limitations with current ways of estimating the predictive effects of future context. We present two alternatives to backward predictability that address issues of collinearity with forward predictability: relative backward predictability, which captures how much more predictable a word is from the future context compared to the past context, and conditional PMI, which quantifies the information gained from the future once the past is taken into account. Unlike backward predictability, conditional PMI measures the unique contribution of the future within that past context, rather than independently of it.

\subsection{Effects of Predictability on Wordform Encoding}
Our first study focused on the effects of predictability on word durations  in naturalistic speech. Previous work has established a robust inverse relationship between predictability and wordform variation: predictability from either direction correlates with reduction in phonetic, acoustic, and articulatory detail. Predictability from the future, in particular, has been observed to have a stronger effect on duration than predictability from the past, although previous work has shown that this effect appears to be modulated by lexical class (\citealt{bell2009predictability} \emph{cf.} \citealt{ranjan2022linguistic}). In our first study, we find that both proposed variants of backward predictability---relative backward predictability and conditional PMI---not only demonstrate this inverse relationship but also emerge as stronger predictors of word duration than forward predictability, in accordance with previous findings. Furthermore, conditional PMI, which captures joint effects of past and future context, explains variance in word durations even alongside relative backward predictability. This finding suggests that such joint effects also shape duration, above and beyond the independent contributions of past and future---which, to our knowledge, has not been demonstrated before.

Previously, \citealt{bell2009predictability} reported asymmetric effects of frequency and contextual predictability on articulatory reduction: whereas frequency and backward predictability emerged as significant predictors of content word duration, forward predictability only affected the duration of high-frequency function words. One possible explanation for the lack of a forward predictability effect on content word durations put forth by Bell and colleagues was that preceding words may be less informative about word choice than following words. Crucially, Bell and colleagues reported no effects of contexts larger than the preceding or following word. 

\begin{table}[t]
    \centering
    \resizebox{.95\textwidth}{!}{%
    \begin{tabular}{l ccc | ccc}
    \toprule
      & \multicolumn{3}{c }{\textbf{Bell et al. (2009)}} & \multicolumn{3}{c}{\textbf{Current Study}} \\
    \cline{2-7}
      & HF Function & MF/LF Function & Content & HF Function & MF/LF Function & Content \\
    \midrule
    Frequency  & \emph{ns} & \emph{ns} & \emph{hs} &  $p < 0.001$ &  $p < 0.001$ & $p < 0.001$\\
    Predictability from the past & \emph{hs} & \emph{ms} & \emph{ns} & $p < 0.001$ & $p < 0.001$ & $p < 0.001$\\
    Predictability from the future & \emph{ns} & \emph{hs} & \emph{hs} & $p < 0.001$ & $p < 0.001$ & $p < 0.001$ \\
    \bottomrule
    \end{tabular}}
    \caption{A comparison of predictability effects for High-Frequency (HF) function, Mid-Frequency (MF)/Low-Frequency (LF) function, and content words observed in Bell et al. (2009) and the current study. Significance as reported by Bell et al. (2009): \emph{ns} = not significant, \emph{ms} = marginally significant, and \emph{hs} = highly significant.}
    \label{tab:summary-comp}
\end{table}

A summary of our key results vis-à-vis \citealt{bell2009predictability}'s findings is presented in Table \ref{tab:summary-comp}. First, we observe significant predictability effects for context windows that span longer than the preceding or following word, consistent with the improved predictive power of neural language models that condition on broader context than bigram models.

Among the three sources of predictability, we find that future context predictability emerges as the strongest predictor of duration, regardless of operationalization. Furthermore, the present study did not replicate the asymmetric predictability effects of past and future context on function versus content word durations, as in \citet{bell2009predictability}. Predictability from either direction led to reduction in function and content words alike, which suggests that the lexical class distinction may reflect an availability-based effect rather than a separate mechanism of access for function versus content words (\emph{cf.} \citealt{gordon2003learning}).

Yet notably, both function and content word durations were more strongly modulated by predictability from the future context, regardless of the operationalization (i.e., relative backward predictability or conditional PMI). \citet{pluymaekers2005articulatory} interpret stronger predictive effects of the following word on the acoustic realization as evidence of continuous and anticipatory articulatory planning: stronger backward associations enable speakers to compress the production of the current word to allocate resources toward planning and articulating the next one. 

A more general interpretation of this effect is that words that are highly predictable from the future context may be produced more quickly, allowing speakers to transition to the production of upcoming words, hence reducing the duration for which these words must be maintained in working memory \citep{ferreira2000effect,wheeldon2006language,slevc2011saying}. Conversely, when upcoming material is difficult to retrieve, speakers may `buy time' for planning by prolonging the duration of the current word \citep{foxtree1997pronouncing,bell2003effects}. 

Although \citet{bell2009predictability} also acknowledge this anticipatory planning interpretation of the backward predictability effect, they frame predictability effects more generally as arising from the need to coordinate between planning and articulation: when the current word is unpredictable given the past or future context, the production system responds to this processing difficulty by slowing down and prolonging the articulation of the current word. Notably, this view treats predictability from the past and future as mirror effects. Empirically, however, future context predictability consistently emerges as the stronger predictor of word duration, not only in \citet{bell2009predictability} but also in the present study, which examined multiple operationalizations of this effect. A possible explanation for the stronger influence of future context on durations is that this effect may capture two sources of difficulty: the predictability of a word given some representation of the future sequence, but also the availability of the future representation itself.

\begin{table}[t]
\centering
\resizebox{\textwidth}{!}{%
\begin{tabular}{l ccc}
\toprule
& \textbf{Relative Backward Predictability} & \textbf{Conditional PMI} & \textbf{Combined} \\
\midrule
\textbf{Word Duration} & Stronger predictor & Weaker predictor & Non-redundant; combined model shows best fit \\
\addlinespace
\textbf{Substitution Choice} & Weaker predictor & Stronger predictor & Redundant; no improvement over conditional PMI \\
\bottomrule
\end{tabular}}
\caption{A comparison of the explanatory power of relative backward predictability, conditional PMI, and both predictors combined across the two dependent variables examined in this study. Non-redundancy indicates that including both predictors yields a significant improvement in model fit over either predictor alone.}
\label{tab:measure-comparison}
\end{table}

\rr{Finally, we address the intriguing finding that relative backward predictability---a measure derived from backward predictability which marginalizes over the past context---emerges as a stronger predictor of word duration than conditional PMI, which by definition captures the informativity of the future context given what has already been produced. Yet that these two measures are non-redundant in explaining duration: including both yields a significant improvement in model fit, suggesting they capture distinct aspects of the relationship between future context and phonetic encoding.}

\rr{We speculate that the marginalization of past context in relative backward predictability yields a decontextualized and aggregated estimate of future context predictability that may provide a better fit to duration, which is a composite phonetic measure that itself integrates multiple bands of planning within the production system. By contrast, conditional PMI likely isolates a more context-specific band of planning that relies on richer representations of past and future context. This may explain why conditional PMI, despite showing an attenuated effect compared to relative backward predictability, still survives as a non-redundant predictor of duration. This interpretation is also consistent with our findings from Study 2, which show that when the dependent variable implicates a more context-sensitive choice i.e., the identity of the substitution error, conditional PMI subsumes relative backward predictability entirely. }

\subsection{The Role of Context in Lexical (Mis-)Selection}
\rr{The core claim of speaker-oriented accounts of probabilistic reduction is that predictability facilitates lexical access, which, in turn, leads to reduced effort in the downstream encoding of the word. Yet the effects of contextual predictability on the \emph{content} of speaker choices remain underexplored in naturalistic speech, owing to the difficulty of inferring the intended lexical meaning that the speakers aim to convey through their choice of lexical representation. In our second study, we examined the effects of lexical, alignment-driven, and contextual factors on word choice in substitution contexts, where the speaker's self-repair serves as an approximation of their production target.}  

\rr{The core contribution of this analysis, however, is to characterize the role of context in lexical (mis-)selection. Prior work offers two complementary insights into how contextual predictability shapes production errors. First, contextual priming can boost the accessibility of competing alternatives, increasing the likelihood that they are mis-selected \citep{bock1985conceptual,rapp2002reason,ferreira2003phonological}. Second, less predictable contexts elicit more disfluencies and are associated with higher error rates \citep{foxtree1997pronouncing,bell2003effects,dammalapati-etal-2019-expectation}. Both, however, concern the \emph{occurrence} of errors. 
To our knowledge, this is the first empirical demonstration that predictability from both past and future context, estimated from the same language model, shapes which word, out of a large lexicon of alternatives, surfaces as a naturalistic substitution error. Notably, our modeling of substitution identity shows that once frequency as well as semantic and form-based alignment are controlled for, past and future context predictability exert opposing influences. Like frequency, higher past context predictability increases the likelihood that a word surfaces as an error, suggesting that more predictable words are also more error-prone. Future context predictability, however, shows the opposite effect: words that are more compatible with the speaker's future context appear \emph{less likely} to surface as errors.}

\rr{In the present study, we adopt a computational-level perspective, focusing on the informational factors that operate during lexical selection. That is, we explicate our findings within a view of lexical selection that is agnostic to assumptions about the nature of lexical representation and stages (or lack thereof) in processing. Instead, we view the lexical (mis-)selections in naturalistic substitutions as resulting from competition between form-, meaning-, and context-based information. While self-repairs may reflect a more optimal prioritization of these informational factors, substitution errors reveal suboptimal weighting of these cues under intrinsic constraints on information processing \citep{bock1982toward,ferreira2002central,futrell2023information}}. 

\rr{A fine-grained categorization of the naturalistic substitutions revealed that roughly 85\% of the substitutions in our analysis could be considered lexico-semantic rather than morphosyntactic or phonological. Consequently, the patterns reported here can be viewed as primarily characterizing this class of errors. Across substitutions, word choice reflected a trade-off between speaker effort and communicative reward: words that surfaced as intrusions tended to be frequent and predictable from the past context, while also reflecting a strong semantic and phonological alignment with the target. Beyond this trade-off, we observed a strong inhibitory effect of future context predictability: words more informative about the speaker's upcoming material were less likely to surface \emph{in error}---an effect qualitatively distinct from the facilitatory influence of frequency and past context predictability.}

\rr{One possible interpretation of this inhibitory effect is that these substitutions arise not from a failure to incorporate future context altogether, but from insufficient integration of future representations into the current selection process. Under this view, while speakers may plan upcoming material in advance \citep{lee2013ways,momma2016timing,sauppe2017word,momma2018unaccusativity,Momma2019BeyondLO}, the degree to which pre-planned representations constrain current lexical selection may vary, as actively maintaining and integrating future plans imposes a further burden on verbal working memory \citep{ferreira2000effect,wagner2010flexibility,slevc2011saying,christiansen2016now}. When this integration is suboptimal, future representations may fail to sufficiently constrain lexical selection, allowing a more accessible but contextually inconsistent alternative to slip through as an error.}

\rr{An alternative yet complementary interpretation is that predictability from the future may serve as a mismatch signal that flags the inadequacy of the error: when the future context is highly informative about the target but inconsistent with the error, the production system may detect this mismatch and initiate repair, thereby reducing the likelihood that the error surfaces as a substitution. While this interpretation does not directly explain the inhibitory effect of future context on substitution choice, it makes a testable prediction---that greater informativity between the future context and the intended word should facilitate faster error resolution---which we leave to future work. Regardless of the mechanistic interpretation, we find that future context predictability qualitatively differs from past context predictability in that it inhibits rather than facilitates mis-selection — suggesting that past and future context exert opposing pressures on lexical choice, reflecting goal-invariant and goal-directed aspects of lexical planning, respectively.} 

\rr{This balance between goal-invariant or goal-driven processing is captured by the idea of \emph{Good-Enough (GN)} choices in production \citep{ferreira2003phonological,koranda2022good,goldberg2022good}. Previously, \citet{koranda2022good} demonstrated a trade-off between message alignment and lexical accessibility in word choice: given a speed-accuracy trade-off and multiple possible alternatives to describe a production target, speakers produced the less precise but more frequently experienced alternatives over the more precise yet less accessible choice. Our study builds on this work in two important ways. Whereas \citet{koranda2022good} contrasted the effects of lexical frequency and semantic alignment in a controlled single-word production paradigm, our study extends the scope to word choice in naturalistic utterances, which enables us to examine the role of context in word choice errors. Second, we adopt a comprehender-oriented notion of communicative reward, which encompasses both semantic alignment and proximity to the phonological form of the target. The latter stems from the assumption that comprehenders can infer the intended form through rational inference \citep{gibson_rational_2013,ryskin2018comprehenders}, particularly in the case of errors that are phonologically similar to the target \citep{upadhye2025spacer}. In other words, errors that are more likely to be correctly inferred by the comprehender may be less costly, and therefore, more communicatively robust compared to those that the interlocutor may find difficult to correct.}

\rr{Under this expanded notion of good-enough production, substitution errors can be understood as reflecting a suboptimal weighting of informational factors under intrinsic processing constraints. More broadly, this framing aligns with a \emph{resource-rational} view of choice behavior, according to which  agents operate under intrinsic cognitive capacity limitations that effectively constrain the information processing that may be used to select the utility-maximizing choice
\citep{simon1956rational,anderson1991adaptive,gigerenzer2008rationality,griffiths2015rational,gershman2015computational,lieder2020resource}. A \emph{rational} response to these constraints is that agents are incentivized to arbitrate between easy-first, heuristic, or automatic, and effortful, goal-driven, or controlled modes of processing, which can lead to choices that reflect a trade-off between utility and information processing costs \citep{schneider1977controlled,Shiffrin1984AutomaticAC,kahneman1984changing,evans2013dual,lai2021policy,futrell2023information}}. 

\subsection{Relationship between Contextual Predictability and Theories of Sentence Planning}
Finally, we discuss the link between contextual predictability and theories of planning in sentence production. It is widely assumed that sentence production proceeds through conceptual, functional, positional, and articulatory-motor stages of processing, with `planning' unfolding at each of these levels of representation \citep{Bock1994LanguageP,ferreira2007grammatical,hickok2012computational}. Under this framework, planning at the conceptual stage involves selecting the relevant semantic, relational, and communicative features of the pre-verbal message; at the functional level, it encompasses both lexical selection and hierarchical structure-building; at the positional level, planning involves morphological encoding and serialization; and at the articulatory-motor level, it involves generating and executing the phonetic plan of the speaker's utterance. These planning processes are assumed to proceed sequentially yet incrementally. As discussed in the preceding section, incrementality in language production emerges as a response to both cognitive constraints and conversational pressures that limit the extent to which speakers can plan their utterance prior to the onset of production (\citealt{clark1986referring,kempen1987incremental,levelt1989speaking,ferreira1996better,ferreira2000effect,ferreira2002incremental,wheeldon2006language,levinson2015timing} \emph{inter alia}). 

However, an outstanding issue in theories of sentence production concerns not just the scope, but also the order in which constituents are planned. A \textbf{strictly linear} view of sentence planning maintains that the order of planning roughly mirrors the order in which words appear in the surface form of the utterance \citep{kempen1987incremental,levelt1989speaking,de199611,griffin2001gaze,gleitman2007give,iwasaki2011incremental,christiansen2016now}. This strategy exemplifies an \emph{economy of effort} principle in planning: by prioritizing the retrieval and encoding of words that are accessible\footnote{As operationalized by various correlates of accessibility such as givenness, imageability, animacy, and subjecthood} \citep{bock1985conceptual,mcdonald1993word,griffin2001gaze,gleitman2007give}, and minimizing the lag between retrieval and production to the extent permissible by word order constraints, speakers can leverage the processing benefits offered by accessibility while reducing demands on working memory \citep{ferreira2000effect,slevc2011saying,christiansen2016now}. 

In comparison, a \textbf{hierarchical} view of sentence planning underscores the role of structural relations in determining the order in which words are retrieved, with verbs occupying a privileged status in planning due to their central role in specifying clause structure \citep{Bock1994LanguageP,ferreira2013syntax,schriefers1998producing,momma2016timing,sauppe2017word,Momma2019BeyondLO}. Under a strong version of this view, words that are linked in a dependency may be retrieved concurrently, regardless of their respective positions in the surface form of the utterance. However, cross-linguistic empirical evidence favors a weaker version of this hypothesis: speakers appear to flexibly arbitrate between highly sequential and hierarchical or look-ahead planning strategies depending on linguistic and meta-linguistic factors such as argument structure, head directionality, word order constraints, and working memory demands \citep{schriefers1998producing,lee2013ways,wheeldon2013lexical,momma2016timing,momma2018unaccusativity,Momma2019BeyondLO,nordlinger2022sentence,kidd2025does}.

Although the present study does not account for how speakers arbitrate between strictly linear and hierarchical planning or generate predictions about the order in which words are planned, these two strategies provide a rationale for the observed effects of past and future context predictability on both word duration and word choice. Strictly linear planning can be viewed as analogous to incremental prediction, wherein the production of previous words facilitates access of contextually predictable continuations, reflecting an accessibility-driven strategy. Hierarchical planning, on the other hand, provides a motivation for the early availability of future material. Since both the scope and content of a speaker's future plans appear to be sensitive to cognitive load, this planning strategy can be viewed as reflecting more effortful or goal-oriented planning. From a resource-rational perspective, this interleaving of planning strategies is consistent with our findings that past and future context predictability appear to reflect goal-invariant and goal-directed pressures on lexical choice, respectively. 

Moreover, this flexibility in planning also provides further motivation for our proposed measure of future context predictability based on conditional PMI$(w_t, C_{>t} \mid C_{<t})$. Although backward predictability quantifies the predictive effect of future context on the present word, this effect is not situated in the context of what the speaker has already produced. Furthermore, by ignoring any direct information-processing dependencies between words in the past and future contexts, the measure assumes that the future context depends only on the current word. An implication of this assumption that is of relevance to theories of sentence planning is that it assumes that words in the future are planned strictly linearly. By preserving dependencies between past and future contexts and providing a contextualized measure of informativity between the speaker's future plans and the current word, conditional PMI can serve as a more interpretable probabilistic variable for modeling the effects of non-incremental planning.

Moreoever, conditional PMI is a \emph{symmetric} quantity that measures how informative the current word and the future context are about each other in the context of past sequence. In earlier discussion, we focused on a backward-looking interpretation, where conditional PMI captures the constraining effect of an already planned future sequence on present choice. However, this quantity is also compatible with a forward-looking or prospective interpretation of planning such as \emph{forward simulation}, which aligns with how planning is widely conceptualized in action planning, optimal control, and reinforcement learning. Forward simulation, broadly construed, involves generating possible future trajectories of actions according to the agent’s policy, conditioned on both the present state (which reflects the outcomes of past actions) and an action sampled at the current time step \citep{sutton1998reinforcement}. Action selection at the current time-step, therefore, is guided by the \emph{future value} of a candidate action i.e., the expected return aggregated over the simulated future trajectories \citep{todorov2009efficient}. In other words, agents choose the current action with the objective of facilitating a future trajectory that maximizes expected utility or communicative value. In the context of sentence planning, this prospective interpretation of conditional PMI can be viewed as speakers choosing words that are informative and facilitate production of any upcoming material. Since the present study assumes the entire future sequence as a single deterministic chunk, our findings are compatible with both interpretations. We leave it to future work to examine how uncertainty in the representation of the future sequence may affect how speakers choose and encode words.

\section{Conclusion}
This work examines the role of past and future context on wordform encoding and word choice in naturalistic language production. The effect of future context or backward predictability, in particular, has long remained understudied compared to the effects of past or forward predictability. Beyond the methodological and statistical confounds linked to language modeling and multi-collinearity, the link between backward predictability and mechanisms of sentence planning is obscured by the assumption that future context is statistically independent of the past context. In this work, we introduce a principled alternative to backward predictability based on the conditional pointwise mutual between the current word and the future sequence conditioned on the previously produced context. Our empirical contributions are two-fold. First, we revisit findings in probabilistic reduction, both as a validation of our proposed alternative and to re-evaluate earlier claims in light of improved measures of contextual predictability. Broadly, results from our first study reveal that predictability from the future exerts a stronger influence on word duration regardless of lexical class, which suggests uncertainty about the future context may further incentivize speakers to slow production. In our second study, we present a model of substitution errors in context, which explicates the unique contributions of availability-based preferences, communicative utility, and context on word choice. Beyond replicating trade-offs between lexical availability and communicative utility previously observed in single-word production, our findings reveal that past and future context exert distinct, and indeed opposing, influences on lexical choice. Overall, our study helps bridge psycholinguistic theories of sentence production with probabilistic and resource-rational approaches to modeling the mechanisms that underlie language production.

\bibliographystyle{apalike}
\bibliography{references}

\pagebreak
\appendix
\section{Data and Code Availability}
Data and analysis scripts used in this study are available at \url{https://osf.io/s2v7t/overview?view_only=5fc92815101c4c3fab933cfd06446e63}. 

\section{Derivation of Conditional PMI from Value-to-go} \label{app:derivation-PMI}
As noted in Section \ref{sec:quantifying-FCP}, conditional PMI is a symmetric quantity that does not commit to a causal planning mechanism. Rather, it captures how informative the current word ($w_t$) and future sequence ($C_{<t} = \{w_{t+1}, \cdots w_{N}\}$) are about each other in the context of what the speaker has already produced ($C_{>t} = \{w_1, \cdots w_{t-1}\}$). Therefore, it is consistent with both a \emph{retrospective} interpretation, where material retrieved in advance of production constrains word choice, and a \emph{prospective} interpretation in which the current word is chosen such that it aids in the production of desired future words. 

Applied to lexical planning, the control-theoretic concept of \emph{value-to-go} can be viewed as the expected value or reward of choosing a word $w_t$, which includes both the value of choosing the current word and expected value of
a future sequence of words ($c = w_{t+1}\cdots w_{N}$):
\begin{equation}
    \mathbb{E}_{c \sim p(. \mid w_t, C_{<t})} [v_{M}(c \mid w_t, C_{<t})] = \sum_{c} v_{M}(c \mid  w_t, C_{<t}) \ \ p(c \mid w_t, C_{<t}) \label{eq:value-to-go}
\end{equation}
Eq.~\ref{eq:value-to-go} denotes the expected future value of choosing a word at time $t$ by computing the expectation or aggregating over all possible future sequences. Here, $v_M$ denotes the \emph{communicative value} of a future ($c$) under the speaker's goal or message $M$ and conditioned on the choice of the current word and the observed past context.

If we want to use this notion of value-to-go as a predictor of speakers' word choices, we are faced with a problem: enumerating all possible future sequences to compute the sum in Eq.~\ref{eq:value-to-go} is highly intractable. Therefore, we make three simplifying assumptions about the possible future context. First, we treat the entire future context as a 
\textbf{one} multi-word chunk or action taken at time-step $t+1$. Furthermore, for simplicity, we consider only the future context actually produced by the speaker $C_{>t}$ instead of computing an expectation over all possible alternative futures. Finally, we hold the communicative value ($v_M$) constant for all choices of $w_t$. In other words, word choice at time $t$ does not change the communicative value of the observed future, which we already treat as fixed with respect to $M$.

These assumptions enable us to simplify Eq.~\ref{eq:value-to-go} to the probability of the future conditioned on a selected word $w_t$ at time $t$ and observed past $C_{<t}$:

\begin{equation}
    p(C_{>t} \mid w_t, C_{<t}) \label{eq:cond-future}
\end{equation}

\noindent Below, we show that $\textrm{conditional PMI}(C_{>t}; w_t 
\mid C_{<t})$ can be derived from Eq. \ref{eq:cond-future} as follows: \\

\noindent Applying Bayes rule, we can rewrite $p(C_{>t} \mid w_t, C_{<t})$ in terms of the bidirectional probability $p(w_t \mid C_{>t}, C_{<t})$, forward word probability $p(w_t \mid C_{<t}$ and the conditional probability of the $p(C_{>t} \mid C_{<t})$:

\begin{equation}
\begin{aligned}
p(C_{>t} \mid w_t, C_{>t}) &= \frac{p(w_t \mid C_{>t}, \ C_{>t}) \ p(C_{>t} \mid C_{<t})}{P(w_t \mid C_{<t})} \\
\end{aligned} \label{eq:pmi-deriv-interim-1}
\end{equation}

\noindent Applying log-transformation to Eq. \ref{eq:pmi-deriv-interim-1}

\begin{equation}
\begin{aligned}
\log p(C_{>t} \mid w_t, C_{>t}) & = \log p(w_t \mid C_{>t}, \ C_{<t}) \ - \log p(w_t \mid C_{<t}) \ + \log p(C_{>t} \mid C_{<t}) \\    
\end{aligned} \label{eq:pmi-deriv-interim-2}
\end{equation}

\noindent Note that $p(c \mid C_{<t})$ is a constant with respect to  $w_t$. Hence, we can further simplify Eq. \ref{eq:pmi-deriv-interim-2} as follows:

\begin{equation}
\begin{aligned}
& \propto \log p(w_t \mid C_{>t}, \ C_{<t}) \ - \log p(w_t \mid C_{<t}) \\
& = \textrm{Conditional PMI}(C_{>t}; w_t \mid C_{<t})
\end{aligned} \label{eq:pmi-deriv-final}
\end{equation}

\pagebreak

\section{Language Model Training and Evaluation}  \label{app: model-training-and-eval}
A GPT-2 \texttt{small} transformer language model was pre-trained using the data augmentation process detailed in Figure \ref{fig:language_modeling}. Instead of using an off-the-shelf Byte-Pair Encoding (BPE) tokenizer \citep{radford2019language}, which relies on subword tokenization based on token frequency, we opt to train a whitespace-based tokenizer on the CANDOR corpus; we choose this tokenization scheme for simplicity of estimating token or word probability and to keep the vocabulary size more tractable (N = 14116). Beyond this, the tokenizer vocabulary also included $\texttt{<eos>}$, $\texttt{<PRE>}$, $\texttt{<SUF>}$, $\texttt{<MID>}$, and $\texttt{<unk>}$ tokens. Below, we provide an algorithmic implementation of the process for augmenting the model training and evaluation datasets. \\

\DontPrintSemicolon
\begin{algorithm}[H]
\caption{Procedure for augmenting training and evaluation datasets for infill estimation}
\KwIn{Corpus $U$}
\KwOut{Augmented corpus $U'$}
\For(\tcp*[f]{Iterate over every utterance in the corpus}){$u \in U$}{
    $N \gets length(u)$
    
    $k \sim \mathcal{U}(1, N)$ \tcp*[r]{Uniformly sample the position of a word}

    $\rho \sim Bernoulli(0.5)$

    \If{prefix precedes suffix in $u'$}{
    $\rho = 0$ \;
    $u' \gets w_1 \cdots w_{k-1} w_{k+1}\cdots w_N w_k$ \tcp*[r]{Prefix precedes suffix in $u'$}
}
\Else{
    $u' \gets w_{k+1} \cdots w_{N} w_{1}\cdots w_{k-1} w_k$ \tcp*[r]{Suffix precedes prefix in $u'$}
}

}
\end{algorithm} 

Model hyperparameters were determined via grid-search and performance was evaluated based on language modeling perplexity. The finalized hyperparameters are summarized in Table ~\ref{table:model-hyperparams}.

\begin{table}[H]
    \centering
    \begin{tabular}{ll}
        \hline
        \textbf{Hyperparameter} & \textbf{Value} \\
        \hline
        Context Window & $1024$ \\
        Initial Learning Rate   & $5 \times 10^{-5}$   \\
        Training Batch Size   & $4$   \\
        Validation Batch Size   & $4$   \\
        Epochs & 10 (with early-stopping) \\
        L2-regularization & 0.01 \\
        \hline
    \end{tabular}
    \caption{Hyperparameters for training GPT-2 \textsc{small} ($124$M parameter) language model.}
    \label{table:model-hyperparams}
\end{table}

Whereas the model was trained on the CANDOR corpus, the duration and substitution utterances were extracted from Switchboard. It bears mentioning that these spontaneous speech corpora were compiled over three decades apart, and the CANDOR corpus comprises conversations over video chat while conversations in Switchboard were over the telephone. Consequently, we anticipate differences across corpora, which may be due to diachronic trends in language use and modality-specific factors such as audio versus visual feedback during conversation. Therefore, to ensure that the model trained on CANDOR generalized to Switchboard, the models were evaluated on a held-out subset of the Switchboard corpus. The infill-trained GPT-2 achieved a perplexity of 16.195 on Switchboard while forward and backward-trained GPT-2 models achieved perplexities of 39.2778 and 38.3563, respectively. The lower perplexity for the infill-trained model suggests that bidirectional contextual information---even when preceding and following contexts are presented outside of canonical order---leads to improved prediction of the word in context.

\begin{figure}[h]
  \centering
  \includegraphics[width=0.62\textwidth]{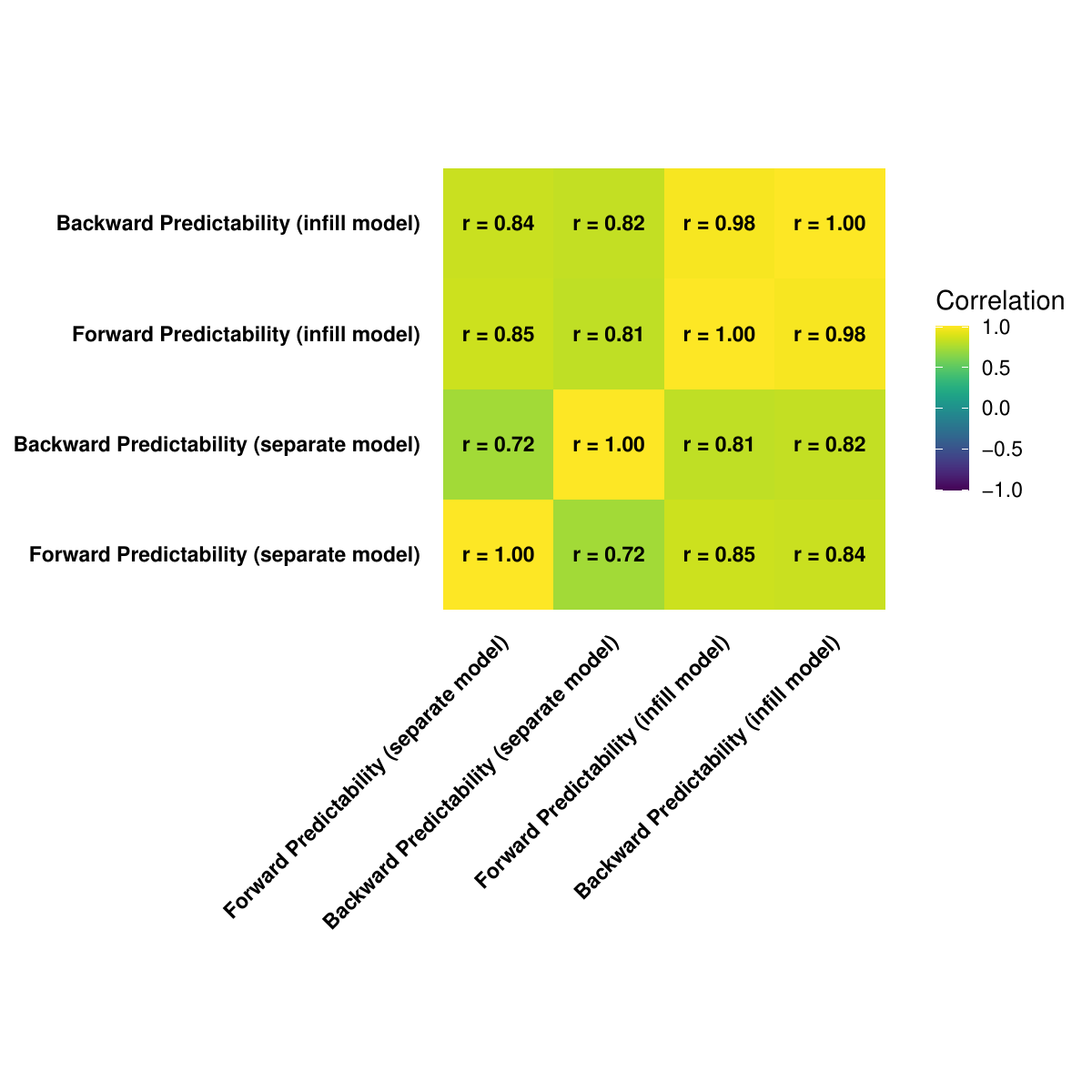}
  \caption{Pairwise Pearson correlations between forward and backward predictability estimates from separate unidirectional and infill-trained models.}
  \label{fig:corr-plot-between-models}
\end{figure}

Forward and backward predictability estimates from both models were found to be highly correlated, suggesting that infill-trained model could reliably estimate autoregressive probabilities in the forward and backward directions. Correlations between these estimates are presented in Figure ~\ref{fig:corr-plot-between-models}

\clearpage

\clearpage
\section{Study 1: Statistical Models and Results} \label{sec:study1-app}

\subsection{Linear regressions for modeling word durations} \label{app:study1-stat-model-specs}
Table \ref{tab:study1-linear-regressions} details the specifications of the mixed effects regression models that were fit to word durations from the Switchboard corpus in Section \ref{sec:study1}.

\begin{table}[b!]
\centering
\begin{tabularx}{\textwidth}{lX}
\toprule
\textbf{Model} & \textbf{Model Equation} \\
\midrule
Model 1 (Baseline) & Duration $\sim$ Unigram Probability + Forward Predictability + Word Length + Speech Rate + Speaker Age + Speaker Sex + (1 | Speaker)\\ \\
Model 1a  & Duration $\sim$ Unigram Probability + Forward Predictability + \textbf{Relative Backward Predictability} + Word Length + Speech Rate + Speaker Age + Speaker Sex + (1 | Speaker)\\ \\
Model 1b  & Duration $\sim$ Unigram Probability + Forward Predictability + \textbf{Conditional PMI} + Predictability + Word Length + Speech Rate + Speaker Age + Speaker Sex + (1 | Speaker)\\ \\
Model 1c (both)  & Duration $\sim$ Unigram Probability + Forward Predictability + Relative \textbf{Backward Predictability} + \textbf{Conditional PMI} + Predictability + Word Length + Speech Rate + Speaker Age + Speaker Sex + (1 | Speaker)\\ \\
Model 1a (predictability * lexical class) & Duration $\sim$ \textbf{Unigram Probability * Lexical Class} + \textbf{Forward Predictability * Lexical Class} + \textbf{Relative Backward Predictability * Lexical Class} + Word Length + Speech Rate + Speaker Age + Speaker Sex + (1 | Speaker) \\ \\
Model 1b (predictability * lexical class) & Duration $\sim$ \textbf{Unigram Probability * Lexical Class} + \textbf{Forward Predictability * Lexical Class} + \textbf{Conditional PMI * Lexical Class} + Word Length + Speech Rate + Speaker Age + Speaker Sex + (1 | Speaker) \\
\bottomrule
\end{tabularx}
\caption{Linear Mixed-Effects Regression models fit using \texttt{lme4}. Incrementally added predictors are denoted in \textbf{bold}.}
\label{tab:study1-linear-regressions}
\end{table}

\pagebreak
\subsection{Regression coefficients for analysis of function and content word durations} \label{app:lexClass-analysis}
Regression coefficients from the lexical class analysis in Section \ref{sec:study1} are summarized in Table \ref{tab:DurModelCoeffs_interactions}.

\begin{table}[h]
\centering
\resizebox{\textwidth}{!}{
  \begin{tabular}{@{}l@{\hskip 8pt}cc@{}}
    \toprule
    & \textbf{Model 1a} & \textbf{Model 1b} \\
    & (with relative backward predictability) & (with conditional PMI) \\
    \addlinespace
    \midrule
    (Intercept) & 150.646(2.91)*** &  149.528(2.88)*** \\
    Unigram Predictability & -17.456(0.26)*** &  -17.721(0.26)*** \\
    Predictability from the past & 2.80(0.25)*** & 3.497(0.25)*** \\
    Predictability from the future & -9.492(0.58)*** & -4.376(0.85)*** \\
    Lexical Class:Content & 1.94(1.56) & -18.136(1.57)*** \\
    Word Length (in syllables) &  82.127(0.37)*** & 81.593(0.37)*** \\
    Speech Rate & -27.618(0.14)*** & -27.943(0.14)*** \\
    Speaker Age &   0.212(0.06)** & 0.218(0.06)*** \\
    Speaker Sex:M &  -8.911(1.37)***  &  -8.198(1.35)*** \\
    \midrule
    Unigram Predictability x Lexical Class:Content & -3.381(0.28)** & -3.886(0.28)** \\
    Predictability from the past x Lexical Class:Content & -6.793(0.28)*** & -6.384(0.28)*** \\
    Predictability from the future x Lexical Class:Content & -27.282(2.82)*** &  -29.67(1.21)*** \\
    \bottomrule
    \end{tabular}
    }
    \caption{Regression coefficients from models with relative backward predictability and conditional PMI as operationalizations of future context predictability. Probabilistic predictors are bolded. Parentheses denote standard error. $p < 0.001$ (***), $p < 0.01$ (**), $p < 0.05$ (*), $p > 0.05$ (\emph{ns}).}
    \label{tab:DurModelCoeffs_interactions}
\end{table}

\pagebreak

\section{Study 1: Additional Analyses} \label{app:study1-add-analyses}
\subsection{Extended Model Comparisons with Backward Predictability and Unconditional PMI} \label{app:study1-model-comps}

In our main analysis, we focused on two operationalizations of future context predictability: relative backward predictability and conditional PMI. Our rationale for using relative backward predictability is that it is effectively decorrelated from forward predictability and  maintains the asymmetry of backward predictability while still yielding a model with equivalent explanatory power. As noted in Section \ref{sec:quantifying-FCP}, unconditional PMI (Eq. \ref{eq:uncond-pmi}) also assumes independent effects of past and future context similar to relative backward predictability. Where these two measures differ, however, is that unconditional PMI is a symmetric measure that quantifies the strength of association between the current word and the future. Below, we replicate the analysis conducted in Section \ref{sec:study1} with standard backward predictability and unconditional PMI to sketch a complete comparison between different operationalizations of future-context predictability. 

\begin{table}[h]
    \centering
    \resizebox{\textwidth}{!}{
    \begin{tabular}{@{}l@{\hskip 8pt}cc@{}}
    \toprule
        & \textbf{Model with backward predictability} & \textbf{Model with unconditional PMI} \\
        \addlinespace
    \midrule
    (Intercept) & 145.981(1.85)*** & 145.981(1.85)*** \\
    \textbf{Unigram Predictability} & -23.676(0.12)*** & -56.252(0.40)*** \\
    \textbf{Predictability from the past} &  29.501(0.39)*** & 29.501(0.39)*** \\
    \textbf{Predictability from the future} &  -32.576(0.40)*** &  -32.576(0.40)*** \\
    Word Length (in syllables) &  88.569(0.38)*** & 88.569(0.38)*** \\
    Speech Rate & -32.767(0.14)*** & -33.163(0.14)*** \\
    Speaker Age & 0.070(0.04) &  0.090(0.04) \\
    Speaker Sex:M &  -12.035(0.86)***  & -12.035(0.86)*** \\
    \bottomrule
    \end{tabular}
    }
    \caption{Regression coefficients from models with backward predictability and unconditional PMI as operationalizations of future context predictability. Probabilistic predictors are bolded. Parentheses denote standard error. $p < 0.001$ (***), $p < 0.01$ (**), $p < 0.05$ (*), $p > 0.05$ (\emph{ns}).}
    \label{tab:DurModelCoeffs-extended}
\end{table}

\begin{figure*}[h!]
    \centering
    \includegraphics[width=0.75\textwidth]{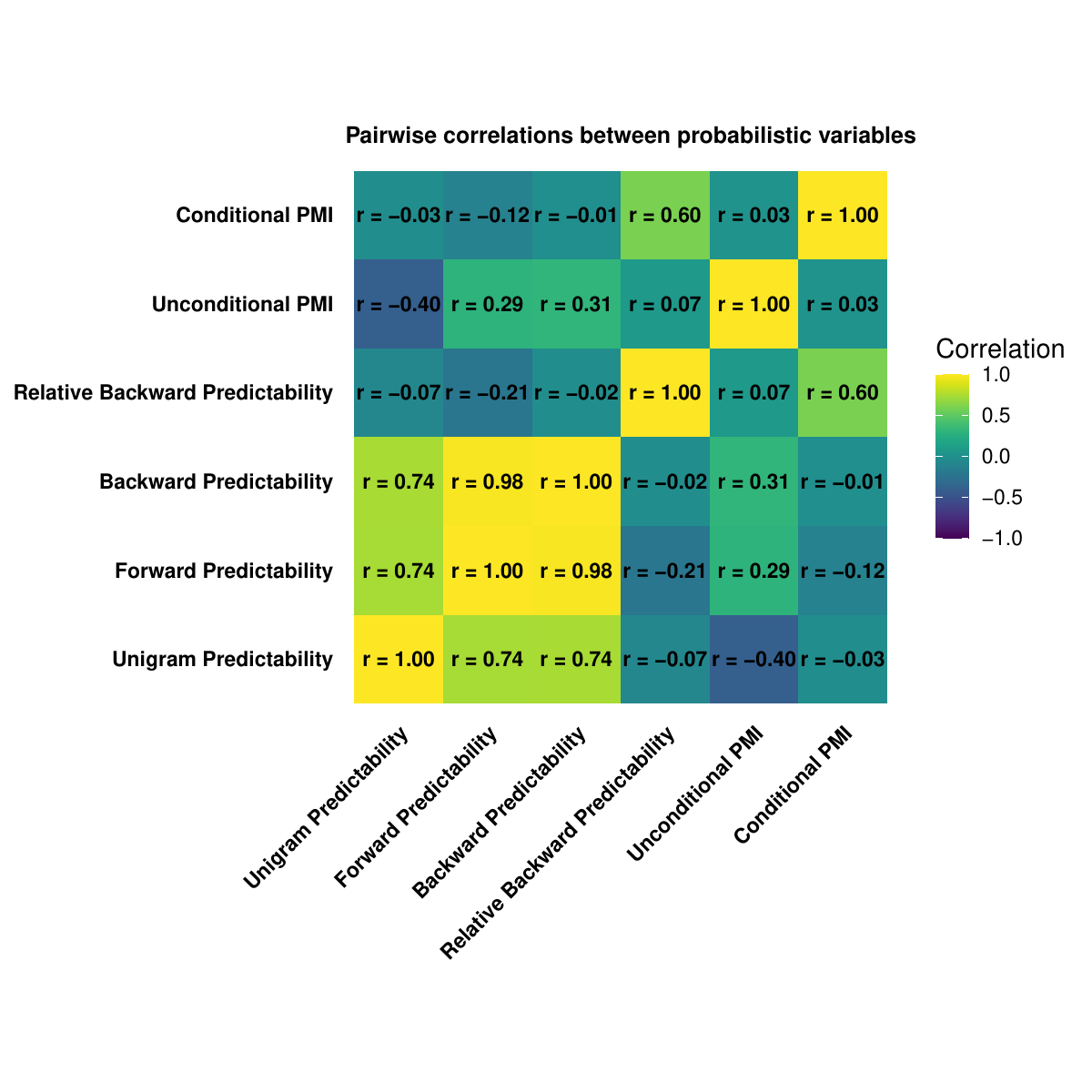}
    \caption{Pairwise Pearson correlation coefficients between all probabilistic variables for all words in the Switchboard corpus} \label{fig:prob-vars-corr}
\end{figure*}

Estimated effect sizes for the probabilistic predictors are presented in Table \ref{tab:DurModelCoeffs-extended}. Model coefficients for backward predictability and unconditional PMI are identical across the two models, with both exhibiting an expected inverse relationship with word duration ($\beta = -32.576, \ SE = 0.40, \ p < 0.001$). Similarly, unigram predictability exhibits an inverse relationship with duration in both models, though this effect is weaker in the backward predictability model ($\beta = -56.252, \ SE = 0.39, \ p < 0.001$) compared to the unconditional PMI model ($\beta = 29.501, \ SE = 0.40, \ p < 0.001$). Intriguingly, the coefficients for forward predictability were also identical across the two models but exhibited an unexpected positive effect on duration ($\beta = 29.501, \ SE = 0.40, \ p < 0.001)$. This unexpected positive relationship between forward predictability and duration not only contradicts previous work, but was also absent in the baseline model and in models with relative backward predictability and conditional PMI.

To determine the cause of this sign flip, we examined the pairwise correlations between all probabilistic predictors (Figure \ref{fig:prob-vars-corr}). First, we find that backward predictability is highly correlated with both unigram predictability ($r = 0.74$) and forward predictability ($r = 0.98$). This suggests that the observed positive effect of forward predictability likely reflects an artifact of the high pairwise correlation between these predictors. Although the model with unconditional PMI addresses the issue of collinearity between backward and unigram predictability, it does not address the issue of co-variance between forward and backward predictability, which may lead to instability in coefficient estimation. 

As discussed in Section \ref{sec:study1-results}, the model with the ``decorrelated'' variant of backward predictability i.e., relative backward predictability not only yielded an effect size of the same magnitude  as backward predictability and unconditional PMI ($\beta = -32.576, \  SE = 0.40, \ p < 0.001$), but also produced an interpretable inverse relationship between duration and forward predictability ($\beta = -3.075, \ SE = 0.11, \ p < 0.001$). Since this measure is effectively uncorrelated with unigram predictability ($r = -0.07$) and exhibits reduced correlation with forward predictability ($r = -0.21$), we argue that relative backward predictability offers a stable alternative to both unmodified backward predictability and unconditional PMI since it mitigates issues of multi-collinearity without affecting explanatory power. 

\pagebreak
\subsection{Comparison with $n$-gram baselines} \label{app: baseline}
In our main analysis, we estimate past or forward predictability and both operationalizations of future context predictability i.e., relative backward predictability and conditional PMI from the a GPT-2 language model trained for infill-based inference. Below we replicate these analysis with forward and backward transitional probabilities estimated from bigram forward and backward models with add-one smoothing.

Table \ref{tab:ngram-lm-comp} provides a comparison of $n$-gram and LM-based estimators. For both $n$-grams and LMs, we include both unmodified backward predictability and relative backward predictability.

\begin{table}[h!]
  \centering
  \resizebox{\textwidth}{!}{
  \begin{tabular}{@{}l@{\hskip 8pt}c cc@{}}
    \toprule
      & $n$-gram baseline & \multicolumn{2}{c}{LM-estimated predictors} \\
      \cmidrule(lr){3-4}
      &  & (with backward predictability) & (with relative backward predictability) \\
    \midrule
    (Intercept) & 135.578(1.81)*** &  145.981(1.85)*** &  145.981(1.85)*** \\
    \textbf{Unigram Predictability} & -23.998(0.13)*** & -23.676(0.12)*** & -23.675(0.12)*** \\
    \textbf{Predictability from the past} & 2.853(0.10) &   29.501(0.39)*** &   -3.075(0.11)*** \\
    \textbf{Predictability from the future} & -4.881(0.096) &   -32.576(0.40)*** &  -32.576(0.40)*** \\
    Word Length (in syllables) & 88.316(0.38)*** & 88.569(0.38)*** &  88.569(0.40)*** \\
    Speech Rate & -33.030(0.14)*** & -32.766(0.14)*** & -32.767(0.14)*** \\
    Speaker Age & 0.077(0.04) &  0.070(0.04) &  0.070(0.04) \\
    Speaker Sex:M & -11.32(0.84) & -12.034(0.86)*** & -12.034(0.86)*** \\
    \bottomrule
  \end{tabular}
  }
  \caption{Regression coefficients from models with bigram-based and LM-based predictors, with relative backward predictability as an operationalization of future context predictability. Probabilistic predictors are bolded. Parentheses denote standard error. $p < 0.001$ (***), $p < 0.01$ (**), $p < 0.05$ (*), $p > 0.05$ (\emph{ns}).}
  \label{tab:ngram-lm-comp}
\end{table}

Under the covariates adopted in the present study, LM-based predictors provide a significantly better fit than $n$-gram predictors ($\deltaLoglik = 1785.441, \chi^{2} = 3574,  p < .001$). This suggests that the relative performance of $n$-gram and LM-based estimators is sensitive to modeling choices. Notably, both estimators exhibit the same sign reversal for forward predictability under unmodified backward predictability, suggesting that this pattern reflects properties of the operationalization rather than the choice of estimator. In contrast, as discussed in Appendix \ref{app:study1-model-comps}, the model with relative backward predictability yields the expected negative effect of forward predictability by reducing collinearity.

\pagebreak

\subsection{Replication of Study 1 with an Intonational Phrase Boundary Control} \label{app:replication-prosody}
Here we replicate the analyses in Section \ref{sec:study1} with the intonation phrase (IP)-final boundary variable. The Switchboard NXT corpus provides prosodic annotations for a smaller subset of the corpus i.e., 73 conversations. Therefore, in this replication, we only consider a subset of the original sample with Tone and Break Indices (ToBI) annotations (27,832 words). While standard prosodic controls include pitch accent and prominence, these are defined at the syllabic-level and rely on prosodic segmentation that does not align one-to-one with the word-level duration measurements provided in the annotations. Because incorporating these variables would require additional assumptions about how prominence or pitch-accent maps onto word-level units, we restrict prosodic controls to speech rate and intonational phrase boundaries.

\rr{\begin{table}[t!]
    \centering
    \resizebox{\textwidth}{!}{
    \begin{tabular}{@{}l@{\hskip 8pt}cc@{}}
    \toprule
      & Model with relative backward predictability & Model with conditional PMI \\
    \midrule
    (Intercept) & 164.13(6.40)*** & 145.957(6.27)*** \\
    Unigram Predictability & -23.167(0.43)*** & -23.638(0.43)*** \\
    Predictability from the past & -3.091(0.42)*** & -2.025 (0.41)*** \\
    Predictability from the future & -32.405(1.45)*** & -38.616(2.24)*** \\
    Word Length (in syllables) &  91.261(1.40)*** & 90.874(1.41)*** \\
    Speech Rate & -36.023(0.53)*** & -36.442(0.53)*** \\
    IP-final Boundary & 4.864(2.04)* & 4.637(1.91)*\\
    Speaker Age & 0.0149(0.14) & 0.065(0.13) \\
    Speaker Sex:M &  -10.77 (2.88)***  & -9.757(2.78)*** \\
    \bottomrule
    \end{tabular}
    }
    \caption{Regression coefficients from models with relative backward predictability and conditional PMI as operationalizations of future context predictability. Probabilistic predictors are bolded. Parentheses denote standard error. $p < 0.001$ (***), $p < 0.01$ (**), $p < 0.05$ (*), $p > 0.05$ (\emph{ns}).}
    \label{tab:DurModelCoeffsPros}
\end{table}}
To account for the robust intonation phrase-final lengthening effect, we extract time-aligned ToBI annotations, and create a binary intonation phrase (IP) boundary variable that encodes whether or not a word occurs at the boundary of a full intonational phrase. Specifically, only ToBI break indices of `4' and `4p' were classified as IP-final. All other break indices were classified as non-final, including 4-, which may denote annotator uncertainty between `3' (intermediate phrase boundary) and '4' (full intonation phrase boundary) \citep{beckman1994tobi}.

We then fit models 1a-1c from Section \ref{sec:study1} with IP-final phrase lengthening as an additional fixed effect. The regression coefficients for models 1a and 1b i.e., models with incrementally added relative backward and conditional PMI predictors, are presented in Table \ref{tab:DurModelCoeffsPros}. Consistent with previous work, we observe a significant effect of IP phrase-final boundary in both model variants i.e., words occurring at the edge intonation phrases are more prone to lengthening. Furthermore, the effects of probabilistic predictors on duration reported in Section \ref{sec:study1} are qualitatively replicated when these prosodic controls are included in the model.

Moreover, replications of the incremental likelihood ratio tests with IP-final phrase boundary included yield a similar pattern of model performance. Specifically, relative backward predictability explains more variance in word duration than conditional PMI ($\deltaLoglik = 97.26, \ \chi^{2} = 194.52, p < .001$), but a model including both predictors provides a further significant improvement in fit ($\deltaLoglik = 13.41, \  \chi^{2} = 30.65, \ p < .001$).

We further replicated the analysis of function (N = 12760) and content (N = 13748) word durations in Section \ref{sec:study1} with IP-final phrase boundary as a prosodic control. Regression coefficients are presented in Table \ref{tab:DurModelCoeffsPros_interactions}.

\begin{table}[H]
    \centering
    \resizebox{\textwidth}{!}{
    \begin{tabular}{@{}l@{\hskip 8pt}cc@{}}
    \toprule
      & Model with relative backward predictability & Model with conditional PMI \\
    \midrule
    (Intercept) & 160.249(7.02)*** &  159.076(6.99)*** \\
    Unigram Predictability & -16.709(0.93)*** & -16.9862(5.67)*** \\
    Predictability from the past &  2.558( 0.93)** &  3.240(0.92)*** \\
    Predictability from the future & -8.273(2.12)*** & -1.910(3.01) \\
    Lexical Class:Content & 4.848(0.93) & -16.304(5.67)** \\
    IP-final Boundary & 4.072(1.96)* & 3.897(1.97)*\\
    Word Length (in syllables) &  83.997(1.34)*** & 83.447(1.35)*** \\
    Speech Rate & -29.332(0.52)*** & -29.707(0.53)*** \\
    Speaker Age & 0.146(0.12) & 0.176(0.12) \\
    Speaker Sex:M &   -9.306(2.61)***  & -8.531(2.56)** \\
    \midrule
    Unigram Predictability x Lexical Class:Content & -2.741(1.03)** & -3.194(1.04)** \\
    Predictability from the past x Lexical Class:Content & -7.377(1.03)*** & -7.080(1.03)*** \\
    Predictability from the future x Lexical Class:Content & -27.206(2.82)*** & -33.696(4.30)*** \\
    \bottomrule
    \end{tabular}
    }
    \caption{Regression coefficients from models with relative backward predictability and conditional PMI as operationalizations of future context predictability. Probabilistic predictors are bolded. Parentheses denote standard error. $p < 0.001$ (***), $p < 0.01$ (**), $p < 0.05$ (*), $p > 0.05$ (\emph{ns}).}
    \label{tab:DurModelCoeffsPros_interactions}
\end{table}

Overall, the interaction between lexical class and probabilistic predictors remains qualitatively unchanged when controlling for IP-final phrase boundaries. This indicates that the main patterns reported in Section \ref{sec:study1} are robust to the inclusion of prosodic controls.

\pagebreak

\section{Study 2: Statistical Models and Results} \label{app:study2-app}

\subsection{Logistic regressions for modeling substitution choice} \label{app:study2-stat-model-specs}
Table \ref{tab:study2-logistic-regressions} provides the logistic regression models that were fit to predict substitution identity in naturalistic contexts in Section \ref{sec:study2}. 
\begin{table}[h!]
\centering
\begin{tabularx}{\textwidth}{lX}
\toprule
\textbf{Model} & \textbf{Model Equation} \\
\midrule
Model 2 (Baseline) & Produced($w_t$) $\sim$ Unigram Probability + Forward Predictability + Noisy Semantic Distance + Noisy Phonetic Distance  \\ \\
Model 2a  & Produced($w_t$) $\sim$ Unigram Probability + Forward Predictability + Noisy Semantic Distance + Noisy Phonetic Distance + \textbf{Relative Backward Predictability} \\ \\
Model 2b  & Produced($w_t$) $\sim$ Unigram Probability + Forward Predictability + Noisy Semantic Distance + Noisy Phonetic Distance  + \textbf{Conditional PMI} \\ \\
Model 2c (Both)  & Produced($w_t$) $\sim$ Unigram Probability + Forward Predictability + Noisy Semantic Distance + Noisy Phonetic Distance + \textbf{Relative Backward Predictability} + \textbf{Conditional PMI} \\ \\
\bottomrule
\end{tabularx}
\caption{Logistic Regression models fit using \texttt{lme4}. Incrementally added predictors are denoted in \textbf{bold}.}
\label{tab:study2-logistic-regressions}
\end{table}

\subsection{Log-odds coefficients from logistic regression models} \label{app:study2-result-tables}
Log-odds coefficients from Models 1a and 1b are summarized in the Table \ref{tab:SubsModelCoeffs}.

\begin{table}[h]
\centering
\resizebox{\textwidth}{!}{
  \begin{tabular}{@{}l@{\hskip 8pt}cc@{}}
    \toprule
    & \textbf{Model 2a} & \textbf{Model 2b} \\
    & (with relative backward predictability) & (with conditional PMI) \\
    \addlinespace
    \midrule
    (Intercept) & 3.523(0.19)*** & 3.256(0.19)*** \\
    Unigram Predictability & 0.849(0.03)*** & 0.849(0.03)*** \\
    Predictability from the past & 0.198(0.03)*** & 0.200(0.03)*** \\
    Predictability from the future & -0.386(0.07)*** & -0.627(0.10)*** \\
    Noisy Semantic Distance & -2.400(0.19)*** & -2.340(0.19)*** \\
    Noisy Phonetic Distance & -0.424(0.04)*** & -0.427(0.04)*** \\
    \bottomrule
  \end{tabular}
}
\caption{Log-odds coefficients from substitution models with relative backward predictability and
conditional PMI as operationalizations of future context predictability. $p < 0.001$ (***), $p < 0.01$ (**),
$p < 0.05$ (*), $p > 0.05$ (\emph{ns}).}
\label{tab:SubsModelCoeffs}
\end{table}

\pagebreak
\section{Model Comparisons with Bayesian Information Criterion} \label{app: bic-model-comp}
In Sections \ref{sec:study1-results} and \ref{sec:study2-results}, we presented model comparisons to evaluate improvements in models' explanatory power when relative backward predictability and conditional PMI were added to the baseline model individually and when they were added together. Since our mixed-effects regression models had a nested structure, and because the goal of our analysis was to identify whether relative backward predictability and conditional PMI contribute redundant or unique sources of variance across both models, we used log-likelihood as the criterion for goodness of fit. To assess whether these findings hold when a more stringent measure of explanatory power is adopted, we also replicate the analysis using the Bayesian Information Criterion (BIC), which additionally penalizes the model for the number of predictors. That is, we may expect BIC to diverge from log-likelihood when comparing (i) the baseline model against models with relative backward predictability or conditional PMI added, and (ii) the model with either predictor alone against the model with both.

\begin{figure}[t]
    \centering
    \begin{subfigure}[t]{0.48\textwidth}
        \centering
        \includegraphics[width=\textwidth]{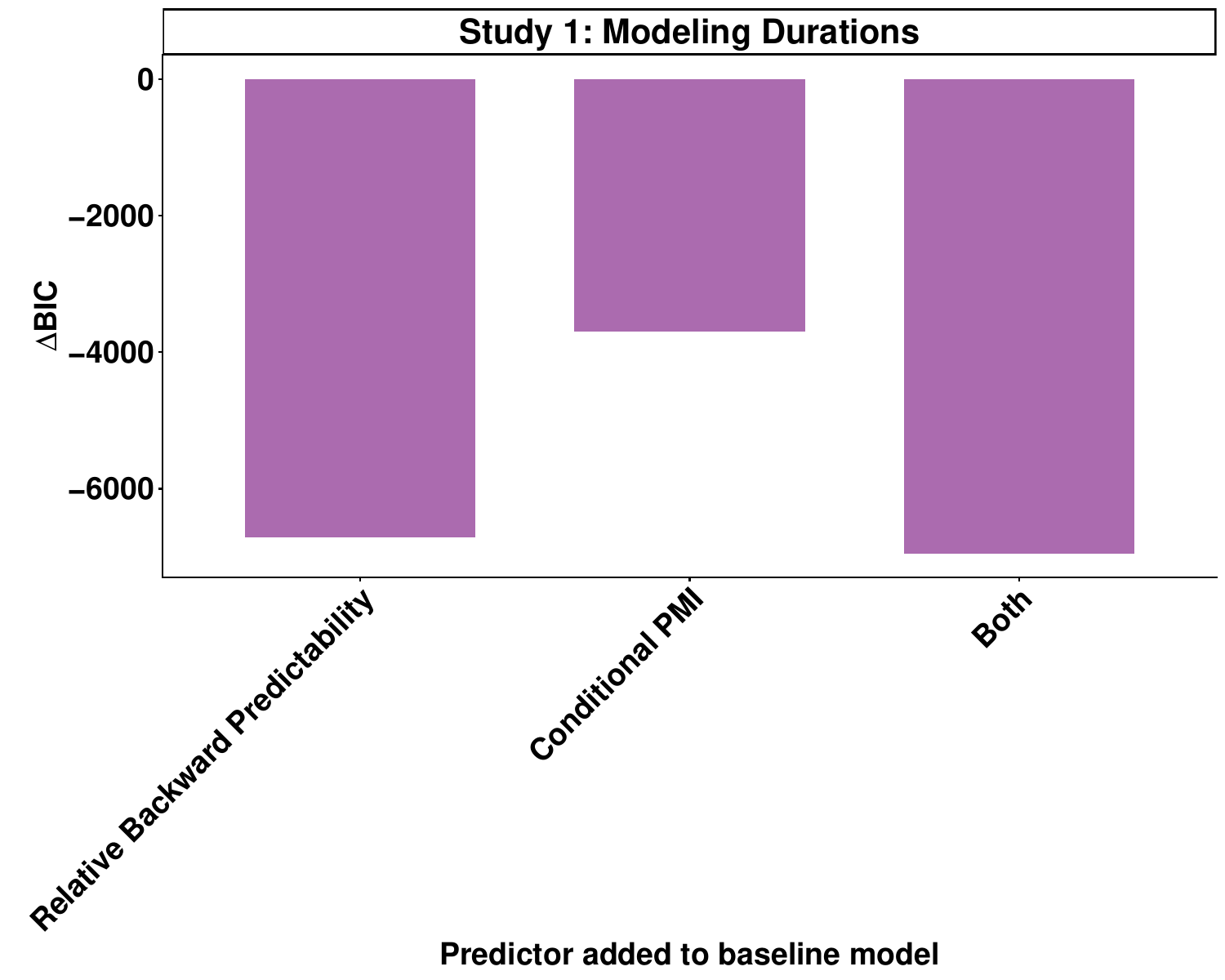}
        \caption{}
        \label{fig:sub1}
    \end{subfigure}
    \hfill
    \begin{subfigure}[t]{0.48\textwidth}
        \centering
        \includegraphics[width=\textwidth]{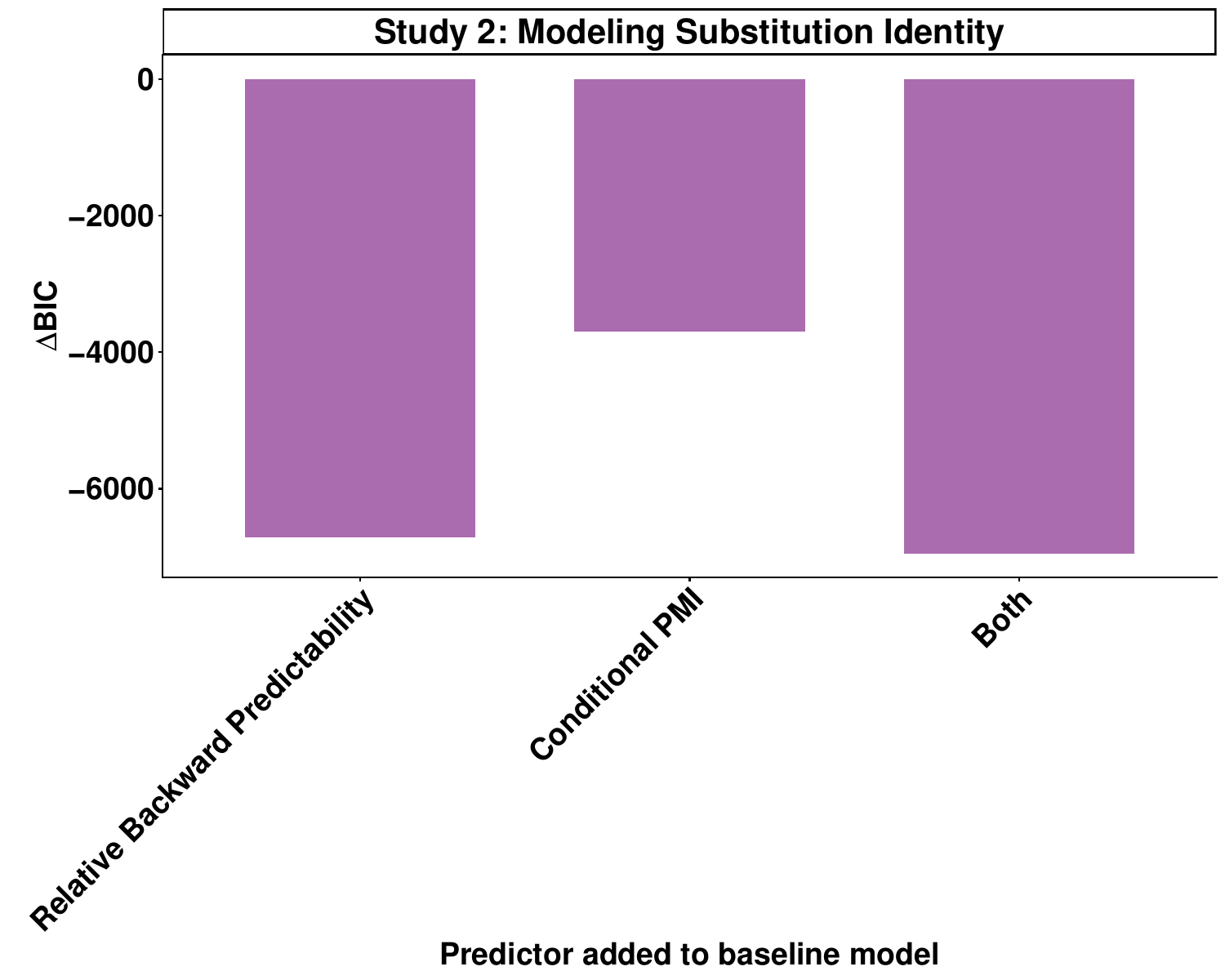}
        \caption{}
        \label{fig:sub2}
    \end{subfigure}
    \caption{Comparisons between models with (i) relative backward predictability only, (ii) conditional PMI only, and (iii) both relative backward predictability and conditional PMI. More negative $\Delta$BIC indicates greater improvement in fit over the baseline model.}
    \label{fig:model-comp-bic}
\end{figure}

BIC-based comparisons for both studies are summarized in Figure \ref{fig:model-comp-bic}. In Study 1, the duration model with relative backward predictability had lower BIC than the model with conditional PMI, indicating a better fit to the data ($\Delta$BIC $= -3012$). Adding conditional PMI to the model with relative backward predictability further reduced the BIC compared to the model with only relative backward predictability; since this difference in BIC $> 10$ ($\Delta$BIC $= -233$), we interpret it as strong evidence that the model with both predictors was a better fit to word durations, even after correcting for model complexity. In Study 2, the substitution model with conditional PMI yielded a lower BIC than the model with relative backward predictability ($\Delta$BIC $= -13.688$). However, when compared to a model with only conditional PMI, the model with both conditional PMI and relative backward predictability had a higher BIC, suggesting that the additional parameter did not improve model performance ($\Delta$BIC $= 14.514$). These results confirm that the findings about the distinct contributions of relative backward predictability and conditional PMI remain robust after adopting a more conservative, complexity-penalizing criterion for model comparison.

\section{Procedure for generating noisy phonetic target} \label{app:algorithms}
\DontPrintSemicolon
\begin{algorithm}[H]
\caption{Algorithm for random perturbation of phonetic features in target representation}
\KwIn{Target word $w_t$}
\KwOut{Noisy target word representation $\hat{w}_t$}
$\Phi \gets \{+, -, 0 \}$ \tcp*[r]{Categorical values of phonetic features}

$N \gets \text{length}(w_t)$

$k \sim \mathcal{U}(1, N)$ \tcp*[r]{Number of phonemes to select}

$p_1, \dots, p_k \overset{\text{i.i.d.}}{\sim} \mathcal{U}(1, N)$ \tcp*[r]{Select positions of the $k$ phonemes}

\For(\tcp*[f]{Iterate over selected $k$ positions}){$p \in \{p_1, \dots, p_k\}$}{
    $f_i \sim \mathcal{U}(1, 22)$ \tcp*[r]{Select feature number to modify}
    
    $\hat{p}[f_i] \sim \mathcal{U}(\Phi \setminus p[f_i])$ \tcp*[r]{Randomly sample an alternative feature value}
}
\end{algorithm}

\section{Preprocessing of a substitution utterance with multiple errors} \label{app:subs-examples}
\begin{enumerate}
\globalitem It depends on whether $\colorbox{red!30}{you}$ whether $\colorbox{green!30}{we}$ figure that we have a defense oriented military or an $\colorbox{red!30}{aggressive}$ $\colorbox{green!30}{aggression}$ oriented military
    \begin{enumerate}
        \item Frame 1: It depends on whether $\colorbox{red!30}{you}$ whether $\colorbox{green!30}{we}$ figure that we have a defense oriented military or an $\colorbox{green!30}{aggression}$ oriented military
        \item Frame 2: It depends on whether $\colorbox{green!30}{we}$ figure that we have a defense oriented military or an $\colorbox{red!30}{aggressive}$ $\colorbox{green!30}{aggression}$ oriented military
    \end{enumerate}
\end{enumerate}

\end{document}